%% file: main.tex
\pdfoutput=1

\documentclass{article} %
\usepackage{paperstyle,times}

\input{math_commands.tex}

\usepackage{hyperref}
\usepackage{url}
\usepackage{amsmath,amssymb}
\usepackage{graphicx}
\usepackage{booktabs}
\usepackage{array}  %
\usepackage{wrapfig}
\usepackage{float}
\usepackage{wrapfig}
\usepackage{subcaption}
\usepackage{xcolor}
\usepackage{afterpage}  %
\newcommand{\rev}[1]{#1}  %
\usepackage{algorithm}
\usepackage{algpseudocode}
\usepackage{placeins}  %

\graphicspath{{figures/}}

\title{Simple Flow Gradient: \\ Fine-Tuning Flow Policies \\ with the Direct \mbox{Q-Gradient}}

\author{Jiahong Tang$^{1,3}$, Jiaju He$^{1,3}$, Zhiyuan Yang$^{1,3}$, Ruyang Han$^{1,3}$, \\
\textbf{Jiahao Cui$^{2,3}$, Haiyue Zhu$^{3}$} \\
\normalfont $^{1}$Nanyang Technological University, Singapore \quad $^{2}$National University of Singapore, Singapore \\
\normalfont $^{3}$Singapore Institute of Manufacturing Technology (SIMTech), \\
\normalfont \phantom{$^{3}$}Agency for Science, Technology and Research (A*STAR), Singapore}  %

\paperfinalcopy %
\begin{document}
\raggedbottom  %

\maketitle
\lhead{}\renewcommand{\headrulewidth}{0pt}  %

\begin{abstract}
Generative policies trained by imitation rarely reach the reliability that deployment demands, and reinforcement learning (RL) offers a way to improve them through feedback from their own interactions.
Off-policy methods reuse past experience for sample efficiency, but applying them to a generative policy raises two concerns identified by prior work: backpropagating the critic's gradient through the sampling chain may be unstable, and an update without an anchor to the data may exploit errors in the critic's value estimates. %
This work examines whether the direct $Q$-gradient, the most elementary such update, can fine-tune a pretrained generative policy without an anchor.
Analyzing direct $Q$-gradient updates on a one-step flow policy, we find that a single sufficiently large update can erase pretrained competence. This motivates bounding the action space change of each episode's update while allowing continued adaptation beyond the pretrained policy. %
\textbf{Simple Flow Gradient (SFG)} applies this bound by projection to a pretrained flow policy updated with the direct $Q$-gradient; the bound is its only addition to the actor-critic recipe, which keeps no anchor term, distillation stage, or auxiliary network. %
Starting from converged behavior cloning policies under matched interaction budgets, SFG reaches a mean task completion score of 0.98 on \rev{seven} manipulation tasks from Robomimic and Franka-Kitchen, and a score of 0.8 more than twice as fast as the fastest baseline. %
Additional experiments show that SFG also applies to multi-step flow policies. Code is available at \url{https://github.com/hxzanyh/sfg}. %
\end{abstract}

\input{sections/introduction}

\input{sections/preliminaries}
\input{sections/analysis}

\input{sections/method}
\input{sections/experiments}

\input{sections/related_work}
\input{sections/conclusion}

\bibliography{references}
\bibliographystyle{paperstyle}

\clearpage  %
\appendix
\raggedbottom  %
\input{sections/appendix_protocol}
\input{sections/appendix_meanflow}
\input{sections/appendix_impl}
\input{sections/appendix_ode}  %
\input{sections/appendix_baselines}
\input{sections/appendix_trust_region}

\end{document}

%% file: math_commands.tex
\usepackage{amsmath,amsfonts,bm}

\def\eqref#1{equation~\ref{#1}}

\def\1{\bm{1}}

\DeclareMathAlphabet{\mathsfit}{\encodingdefault}{\sfdefault}{m}{sl}
\SetMathAlphabet{\mathsfit}{bold}{\encodingdefault}{\sfdefault}{bx}{n}

\newcommand{\E}{\mathbb{E}}

%% file: sections/introduction.tex
\section{Introduction}
\label{sec:intro}

\begin{wrapfigure}{r}{0.48\linewidth}
\vspace{-\baselineskip}
\centering
\includegraphics[width=\linewidth]{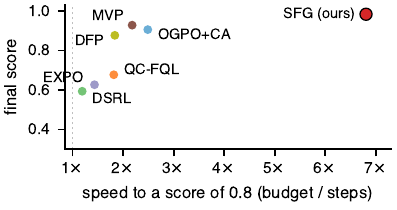}  %
\caption{SFG fine-tunes a pretrained flow policy with the direct Q-gradient, no anchor, and one bound on the step of each episode. How many times sooner than the 1.4M budget the task completion score reaches 0.8 (1$\times$ if never), against the final score, \rev{seven-task} mean of Table~\ref{tab:main}.}
\label{fig:teaser}
\vspace{-\baselineskip}
\end{wrapfigure}
Generative policies, which model actions with diffusion~\citep{chi2023diffusion,ze2024dp3,octo2024} or flow matching~\citep{black2024pi0,intelligence2025pi05,nvidia2025gr00t}, have been widely adopted for robot manipulation and are typically trained by imitation learning. However expressive, such policies often fall short of the reliability that deployment demands~\citep{luo2025hilserl,intelligence2025pistar06}. The learned policy may fit the demonstrations imperfectly, and small action errors compound over long horizons, driving the policy into states that the demonstrations do not cover~\citep{ross2010efficient,ross2011reduction}; its performance degrades further under changes in the scene or the task conditions~\citep{xie2024decomposing,pumacay2024colosseum}. Reinforcement learning (RL) offers a way to improve these policies through feedback from their own interactions. How, then, should it be applied to a pretrained generative policy? %

We focus on off-policy actor--critic methods~\citep{lillicrap2016ddpg,fujimoto2018td3,haarnoja2018sac} rather than on-policy approaches~\citep{ren2024dppo,zhang2025reinflow}, as they can reuse past experience to improve sample efficiency. Applying them to a generative policy raises two concerns: backpropagating the critic's gradient through the sampling chain may be unstable~\citep{patil2026ogpo,zhang2026sacflow,doo2026qflow}, and a policy update without an anchor to the data may exploit errors in the critic's value estimates~\citep{wang2023diffusionql,patil2026ogpo,dong2026trqam}. These concerns have shaped the update designs of existing methods. Many update the policy indirectly, by steering, editing, or filtering the samples of a base policy~\citep{wagenmaker2025dsrl,dong2025expo,li2025qc}, by regressing onto critic-optimized actions~\citep{mark2024parl,ziakas2026fmq}, by distilling a one-step actor~\citep{park2025fql}, or by optimizing a surrogate objective over the denoising chain~\citep{patil2026ogpo,doo2026qflow,wang2026qvgm,dong2026trqam}. Even those that backpropagate the critic's gradient directly keep an anchor to the demonstrations or to the pretrained policy~\citep{wang2023diffusionql,zhang2026sacflow}. %
These designs leave aside the simplest alternative: the \emph{direct $Q$-gradient} without an actor anchor. Can this update effectively fine-tune a pretrained generative policy? %

To answer this question, we analyze the direct update on a one-step flow policy, removing the sampling chain and its associated backpropagation difficulties. We find that direct critic optimization can effectively improve pretrained flow policies without an anchor. A single sufficiently large update along the critic's gradient can erase the pretrained competence, which motivates measuring and bounding the action space change of each update. We therefore propose \textbf{Simple Flow Gradient (SFG)}, which updates the actor with the direct $Q$-gradient and bounds the action space change of each episode's update. The bound lets the policy improve progressively without tying it to its pretrained behavior, and it is the only addition to the actor-critic recipe, which keeps no anchor term, distillation stage, or auxiliary network. In our experiments, every method starts from a behavior cloning policy trained to convergence and runs under a matched interaction budget. Across \rev{seven} manipulation tasks from Robomimic and Franka-Kitchen, SFG achieves a mean task completion score of 0.98, compared with \rev{0.93} for the best-performing baseline. It reaches a score of 0.8 more than twice as fast as the fastest baseline (Figure~\ref{fig:teaser}). SFG also improves multi-step flow policies with the same update, using full or truncated backpropagation through the sampling chain (Section~\ref{sec:exp_ablation}, Appendix~\ref{app:ode}). %

%% file: sections/preliminaries.tex
\section{Preliminaries}
\label{sec:prelim}

\subsection{Flow Matching and MeanFlow}
\label{sec:prelim_meanflow}

Diffusion and flow-matching policies represent a policy $\pi(a \mid s)$ implicitly, through a generative process that transforms Gaussian noise into actions. In flow matching~\citep{lipman2023flow,liu2023rectified}, a demonstration pair $(s,a)$ and a noise sample $\epsilon \sim \mathcal{N}(0,I)$ define the linear interpolation $z_t = (1-t)\,a + t\,\epsilon$ for $t \in [0,1]$, whose conditional velocity is $\epsilon - a$. A state-conditioned velocity field is trained by regression onto this target,
\begin{equation}
\mathcal{L}_{\mathrm{FM}}(\theta) = \E_{(s,a) \sim \mathcal{D},\, \epsilon \sim \mathcal{N}(0,I),\, t \sim \mathcal{U}[0,1]} \big\| v_\theta(z_t, t \mid s) - (\epsilon - a) \big\|^2 .
\label{eq:fm}
\end{equation}
At inference, an action is generated by integrating $\mathrm{d}z_t/\mathrm{d}t = v_\theta(z_t,t \mid s)$ from $z_1 = \epsilon$ to $t=0$, typically with $K$ Euler steps $z_{t-\Delta} = z_t - \Delta\, v_\theta(z_t, t \mid s)$. Backpropagating the critic's gradient through this sampler requires differentiation through the $K$ iterative steps, which prior work has identified as a potential source of instability~\citep{park2025fql,zhang2026sacflow}.

MeanFlow~\citep{geng2025meanflow} replaces the velocity by the \emph{average} velocity over an interval, learned by a network $u_\theta(z_t, r, t \mid s)$ from the same data without a teacher or distillation, so that $z_t$ is carried to $z_r$ in one evaluation (Appendix~\ref{app:meanflow}). With $r = 0$ and $t = 1$ the policy is the one-step generator $a = f_\theta(s,\epsilon) = \epsilon - u_\theta(\epsilon, 0, 1 \mid s)$, $\epsilon \sim \mathcal{N}(0,I)$: a single network evaluation produces the action, and $f_\theta$ is an ordinary differentiable function of $\theta$. The output is clipped to the action box $\mathcal{A} = [-1,1]^d$ inside the decode, and we write $f_\theta$ for the clipped map throughout: it is what is executed, what the critic is trained on and evaluated at, and what the probe of Section~\ref{sec:analysis} measures; the clip enters the gradient only through its exact derivative, which is zero on a saturated coordinate. %

\subsection{Off-Policy Actor--Critic Reinforcement Learning}
\label{sec:prelim_rl}

We consider a Markov decision process $\mathcal{M} = (\mathcal{S}, \mathcal{A}, P, r, \gamma, \rho_0)$ with state space $\mathcal{S}$, action space $\mathcal{A} = [-1,1]^{d}$, transition kernel $P$, reward function $r$, discount factor $\gamma \in [0,1)$, and initial state distribution $\rho_0$. The goal is to find a policy that maximizes the expected discounted return $J(\pi) = \E_{\pi}\big[\sum_{t \ge 0} \gamma^{t} r(s_t,a_t)\big]$. Following common practice for generative robot policies~\citep{chi2023diffusion,black2024pi0} and action-chunked reinforcement learning~\citep{li2025qc}, the policy predicts a chunk of $H$ consecutive actions every $H$ environment steps and executes it open loop. We write a chunk as $a \in \mathcal{A}^{H}$, call it an action, and let $Q^{\pi}(s_t,a_t)$ be the expected discounted return of executing $a_t$ from $s_t$ and following $\pi$ thereafter.

Off-policy actor--critic methods~\citep{lillicrap2016ddpg,fujimoto2018td3,haarnoja2018sac} approximate $Q^{\pi}$ with a critic $Q_\phi$ trained on transitions from a replay buffer $\mathcal{D}$ by minimizing the temporal-difference error
\begin{equation}
\mathcal{L}_Q(\phi) = \E_{(s_t,a_t,s_{t+H}) \sim \mathcal{D}} \Big[ \big( Q_\phi(s_t,a_t) - r^{H}_t - \gamma^{H}\, Q_{\bar\phi}(s_{t+H}, a') \big)^2 \Big], \qquad r^{H}_t = \textstyle\sum_{k=0}^{H-1} \gamma^{k} r_{t+k},
\label{eq:td}
\end{equation}
where $\bar\phi$ denotes slowly updated target parameters and $a' \sim \pi(\cdot \mid s_{t+H})$. When the policy is a differentiable generator $a = f_\theta(s,z)$ of the state and a noise variable $z \sim p(z)$ (Section~\ref{sec:prelim_meanflow}), it can be improved by ascending the critic along its action gradient, as in the deterministic policy gradient and its reparameterized variants~\citep{silver2014dpg,haarnoja2018sac}. The actor minimizes
\begin{equation}
\mathcal{L}_\pi(\theta) = -\E_{s \sim \mathcal{D},\, z \sim p(z)} \Big[ \min_{i \in \mathcal{M}} Q_{\phi_i}\big(s, f_\theta(s,z)\big) \Big] + \lambda\, \mathcal{R}(\theta; \mathcal{D}),
\label{eq:actor}
\end{equation}
where the critic is an ensemble of $N$ networks and $\mathcal{M} \subset \{1, \dots, N\}$ a random subset of $M$ of them, drawn afresh for every update; the same subsampled minimum over the target networks replaces $Q_{\bar\phi}$ in~\eqref{eq:td}. The subsampled minimum in the target is the aggregation of REDQ~\citep{chen2021redq}, which RLPD adopts. REDQ and RLPD then update the actor against the ensemble mean; we use the subsampled minimum in the actor loss as well, a choice of this paper, and it is the only form of pessimism in the update. %
The term $\mathcal{R}$ pulls the policy toward the data or toward the pretrained policy, as a generative behavior cloning loss on demonstrations, a diffusion loss or the flow matching loss~\eqref{eq:fm}~\citep{wang2023diffusionql,nguyen2026ofql,wang2026meanflowql}, a distillation loss~\citep{park2025fql}, a proximity penalty~\citep{zhang2026sacflow}, a loss on successful online episodes~\citep{patil2026ogpo}, or a hard radius around the pretrained policy~\citep{ziakas2026fmq,dong2026trqam}; we call any such term an \emph{anchor} and an update with $\lambda = 0$ \emph{anchor-free}.

\subsection{Online RL with Offline Data}
\label{sec:prelim_rlpd}

When demonstrations are available, RLPD~\citep{ball2023rlpd} shows that an off-policy actor--critic can exploit them without offline pretraining and without any constraint on the policy: each batch is drawn half from the demonstrations and half from online transitions, and the critic is a layer-normalized ensemble with a pessimistic target~\citep{chen2021redq}. RLPD trains a Gaussian policy from scratch on top of this critic. We adopt its replay and critic unchanged and replace the actor with a pretrained generative policy. The demonstrations stay in the replay buffer, and the actor draws its states from the same mixed batches as the critic, so demonstration states do enter the actor loss; what is absent is any term that pulls the generated actions toward the demonstrated ones or toward the pretrained policy. %

%% file: sections/analysis.tex
\section{Where the Direct Update Fails}
\label{sec:analysis}

\subsection{Problem Setting}
\label{sec:analysis_setting}

For a cleaner analysis, we use a one-step generator (Section~\ref{sec:prelim_meanflow}) to remove the iterative sampling chain, avoiding the BPTT difficulties highlighted in prior work~\citep{zhang2026sacflow,doo2026qflow}.
What remains to test is the anchor, which one-step generators have kept as well~\citep{nguyen2026ofql,wang2026meanflowql}. %
Multi-step sampling and alternative backpropagation schemes are evaluated in Section~\ref{sec:exp_ablation}. %

All runs in this section start from one generator $\theta_0$ on \texttt{tool\_hang}, a state-based manipulation task from Robomimic~\citep{mandlekar2021robomimic}, pretrained to convergence with the MeanFlow objective; unless otherwise stated, they use the critic and replay scheme of Section~\ref{sec:prelim_rlpd}.

\subsection{The Direct Update Destroys What Pretraining Built}
\label{sec:analysis_failure}

\begin{figure}[t]
\centering
\begin{subfigure}[t]{0.30\linewidth}
\centering
\includegraphics[width=\linewidth]{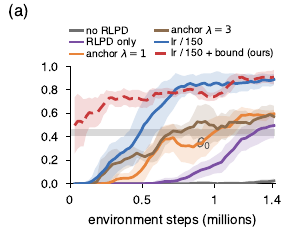}
\phantomcaption
\label{fig:failure}
\end{subfigure}\hfill
\begin{subfigure}[t]{0.30\linewidth}
\centering
\includegraphics[width=\linewidth]{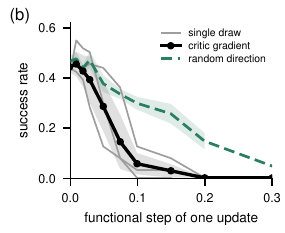}
\phantomcaption
\label{fig:step}
\end{subfigure}\hfill
\begin{subfigure}[t]{0.30\linewidth}
\centering
\includegraphics[width=\linewidth]{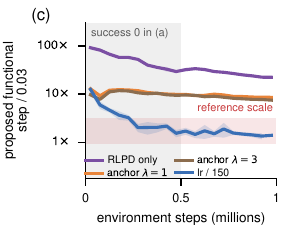}
\phantomcaption
\label{fig:raw}
\end{subfigure}
\caption{\texttt{tool\_hang}, every run from one pretrained $\theta_0$. (a) Success of direct-update variants, with SFG (dashed) as a method reference. The smaller-learning-rate runs use $\tau = 0.005$ without the bound and $\tau = 0.05$ for SFG; Table~\ref{tab:ablate} isolates the bound at matched $\tau$. The gray band is $\theta_0$. (b) Success of $\theta_0$ moved along frozen gradients of a good critic, single draws (gray) and their mean (black), and along random parameter directions at the same functional step (dashed). (c) Proposed functional steps before projection for the settings of (a), normalized by 0.03, a reference scale from (b) (red band); the gray interval is the phase of zero success of the unanchored runs.} %
\label{fig:failure_step}
\end{figure}

We fine-tune with the actor loss~\eqref{eq:actor} at an actor learning rate of $4.5\times10^{-5}$ from a prior that succeeds about half of the time, and remove one component at a time (Figure~\ref{fig:failure}).
With RLPD and a behavior cloning anchor toward the demonstrated actions (Appendix~\ref{app:protocol}), success drops to zero within the first evaluations, remains there for a fifth to a third of the budget, and then recovers to 50 to 60\%.
Without the anchor, the zero-success phase stretches to roughly half of the budget, and the endpoint changes by less than the seed-to-seed spread (Appendix~\ref{app:protocol}).
Without RLPD as well, so that the demonstrations are kept out of every batch, the critic's and the actor's, success stays at zero for the whole budget.

\textbf{Finding 1.} With demonstration replay, the direct update loses the pretrained competence before relearning, both with and without an actor anchor, and ends near the level it started from; the anchor shortens the zero-success phase but does not remove it.

\subsection{A Single Step Is Already Too Large}
\label{sec:analysis_step}

\textit{How large a step along the critic gradient can $\theta_0$ tolerate?}
To examine the effect of step size with a fixed critic, we take a well-trained critic (Appendix~\ref{app:protocol}), freeze it, and compute the direct update once at $\theta_0$.
We then move $\theta_0$ along that gradient by a chosen distance and evaluate, with no training in between (Figure~\ref{fig:step}).
A step of a hundredth of a percent of the parameter norm already halves its success rate, although the successful run ends hundreds of times further from $\theta_0$.
This motivates measuring a policy update by its effect on the generated actions: for a change of the actor from $\theta$ to $\theta'$ we define the \emph{functional step}
\begin{equation}
d(\theta, \theta') = \Big( \frac{1}{|\mathcal{B}|} \sum_{(s,\epsilon) \in \mathcal{B}} \big\| f_{\theta'}(s,\epsilon) - f_{\theta}(s,\epsilon) \big\|^2 \Big)^{1/2},
\label{eq:fstep}
\end{equation}
the root-mean-square change of the generated action chunk over a fixed probe bank $\mathcal{B}$ of states and noise draws, sampled once from the demonstrations and held fixed for the whole run (Appendix~\ref{app:protocol}). %
Small functional steps preserve the pretrained policy's performance, whereas larger steps lead to a sharp drop in success (Figure~\ref{fig:step}). %
At matched functional step sizes, random directions cause a smaller loss of success than critic-gradient directions, although performance declines along both as the step grows. A random direction needs about a hundred times the parameter displacement of the gradient to move the actions by the same functional step. %
In the runs of Figure~\ref{fig:failure}, the optimizer proposes much larger functional steps, particularly in the first episodes, when the pretrained competence is lost (Figure~\ref{fig:raw}). The anchor reduces these steps severalfold, but the initial collapse persists.

\textbf{Finding 2.} Small functional steps preserve pretrained performance in the probe, while the updates of direct fine-tuning induce much larger changes in the generated actions.

A first remedy is a smaller learning rate.
With a 150-fold smaller learning rate, success still collapses at the start, for the first 150 to 250 thousand steps, but then recovers faster and further than any anchored run, to 85\% at one million steps against 60\% (Figure~\ref{fig:failure}).
Even at a fixed learning rate, the proposed functional steps vary substantially over training: at the small rate they are eight times larger early in training than later (Figure~\ref{fig:raw}). %
This variation motivates bounding the functional step directly, as developed in Section~\ref{sec:method}. %

%% file: sections/method.tex
\section{Method: Bounding the Functional Step}
\label{sec:method}

\subsection{Functional Step Projection}
\label{sec:method_cap}

Motivated by Section~\ref{sec:analysis}, we impose a bound $\delta$ on the functional step~\eqref{eq:fstep} of each episode's update and enforce it through \emph{functional step projection}. %
Two copies of the actor take part. Within each episode, the \emph{proposal} actor $\tilde\theta$ collects the data and receives the gradient updates, while the \emph{committed} actor $\theta$ stays fixed as the reference. %
At the episode boundary, the accumulated update $\Delta\theta = \tilde\theta - \theta$ is rescaled to satisfy the functional step bound: the committed actor becomes
\begin{equation}
\theta^{+} = \theta + \alpha\,\Delta\theta, \qquad d(\theta, \theta^{+}) \le \delta,
\label{eq:commit}
\end{equation}
with $\alpha = 1$ when the proposal already satisfies the bound, and otherwise an $\alpha \in (0,1)$ satisfying $d(\theta, \theta + \alpha\,\Delta\theta) = \delta$, found by a scalar root search on the probe bank. In practice, the projection is solved approximately, to a numerical tolerance (Appendix~\ref{app:impl}). %
The bound limits the change of the committed actor from one episode to the next on the probe bank; the proposal that collects the episode is not bounded within it. The reference advances with the committed actor, allowing continued adaptation beyond the pretrained policy. In action space the bound shortens the path the policy travels by more than an order of magnitude without keeping it near $\theta_0$, which is what an anchor does (Appendix~\ref{app:drift}). %
The proposal is then reset to $\theta^{+}$, with the optimizer state kept, and the next episode begins. %
We use an actor learning rate of $3\times10^{-7}$ and $\delta = 0.01$ on every task.

\subsection{The Full Algorithm}
\label{sec:method_alg}

\emph{Simple Flow Gradient} (SFG) adopts the critic and mixed replay of RLPD (Section~\ref{sec:prelim_rlpd}), retaining access to demonstration experience when online success is scarce~\citep{vecerik2017ddpgfd,nair2018overcoming}.
The actor is a pretrained flow policy with its observation encoder frozen, trained on the direct update, \eqref{eq:actor} with $\lambda = 0$.
Its loss is normalized by the batch mean of $|Q|$ (Appendix~\ref{app:impl}), and its updates are applied through the functional step projection of Section~\ref{sec:method_cap}.
The generator may equally be a $K$-step sampler: the functional step is then measured on its output, and nothing else in Algorithm~\ref{alg:direct} changes.
Algorithm~\ref{alg:direct} gives the full procedure and Appendix~\ref{app:impl} the implementation details and hyperparameters.

\begin{algorithm}[!ht]
\small %
\caption{Simple Flow Gradient (SFG): anchor-free fine-tuning with functional step projection}
\label{alg:direct}
\begin{algorithmic}[1]
\Require generator $\theta_0$; demonstrations $\mathcal{D}_{\mathrm{demo}}$; random critic $\phi$; bound $\delta$; probe bank $\mathcal{B}$
\State $\theta, \tilde\theta \gets \theta_0$; $\bar\phi \gets \phi$; $\mathcal{D}_{\mathrm{on}} \gets W$ steps with $\theta_0$, then $W$ critic updates (lines 7--8), $\tilde\theta$ fixed \Comment{warm-up}
\For{each episode}
    \State $\tilde\theta \gets \theta$ \Comment{the proposal starts from the committed actor}
    \For{each action chunk, every $H$ environment steps}
        \State $a \gets f_{\tilde\theta}(s,\epsilon)$, $\epsilon \sim \mathcal{N}(0,I)$ \Comment{a new chunk}
        \For{each of the $H$ actions of $a$, after executing it} \Comment{one critic and one actor update per step}
            \State sample $H$-step segments $(s, a, r^{H}, s_{+H})$, half $\mathcal{D}_{\mathrm{demo}}$ and half $\mathcal{D}_{\mathrm{on}}$; draw head subsets $\mathcal{M}, \mathcal{M}'$
            \State update $\phi$ on~\eqref{eq:td} with target $\min_{i \in \mathcal{M}} Q_{\bar\phi_i}$, $a' = f_{\tilde\theta}(s_{+H},\epsilon')$; \quad $\bar\phi \gets \tau\phi + (1-\tau)\bar\phi$
            \State update $\tilde\theta$ by one optimizer step on $-\E_{s,\epsilon}\big[\min_{i \in \mathcal{M}'} Q_{\phi_i}\big(s, f_{\tilde\theta}(s,\epsilon)\big)\big] \big/ \mathrm{sg}\,\E|Q|$
        \EndFor
    \EndFor
    \State add the episode's transitions to $\mathcal{D}_{\mathrm{on}}$
    \State $\theta \gets \theta + \alpha\,(\tilde\theta - \theta)$, with $\alpha = 1$ if $d(\theta, \tilde\theta) \le \delta$, else the root of $d\big(\theta, \theta + \alpha(\tilde\theta - \theta)\big) = \delta$
\EndFor
\end{algorithmic}
\end{algorithm}
\FloatBarrier

%% file: sections/experiments.tex
\section{Experiments}
\label{sec:experiments}

We evaluate SFG's final performance and sample efficiency against existing fine-tuning methods, then examine the roles of its components and sampler.

\subsection{Setup}
\label{sec:exp_setup}

\paragraph{Tasks.}
We use \rev{eight} sparse-reward manipulation tasks. From Robomimic~\citep{mandlekar2021robomimic}: \rev{\texttt{lift} and \texttt{can}, short-horizon lifting and pick-and-place;} \texttt{square}, a medium-horizon insertion; \texttt{tool\_hang}, a long-horizon multi-step insertion that demands precision; and \texttt{transport}, a bimanual long-horizon transfer; \texttt{square} and \texttt{transport} use the multi-human demonstrations, \texttt{tool\_hang}\rev{, \texttt{can}, and \texttt{lift}} the proficient-human ones.
From Franka-Kitchen~\citep{gupta2019kitchen}: \texttt{complete}, whose demonstrations perform all four subtasks in order; \texttt{mixed}, where the order varies; and \texttt{partial}, where many demonstrations stop early.
Reward is one on success. We report a \emph{task completion score}: the success rate on the Robomimic tasks and the fraction of the four subtasks completed on the kitchen tasks; means over tasks average the two.

\afterpage{%
\begin{figure}[t]
\centering
\begin{subfigure}[t]{0.24\linewidth}\centering\includegraphics[width=\linewidth]{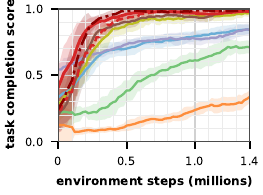}\caption{Robomimic-square}\end{subfigure}\hspace{0.01\linewidth}%
\begin{subfigure}[t]{0.24\linewidth}\centering\includegraphics[width=\linewidth]{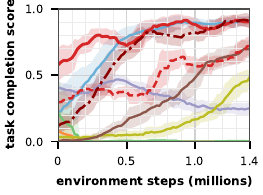}\caption{Robomimic-tool\_hang}\end{subfigure}\hspace{0.01\linewidth}%
\begin{subfigure}[t]{0.24\linewidth}\centering\includegraphics[width=\linewidth]{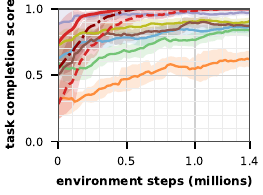}\caption{\rev{Robomimic-can}}\end{subfigure}\hspace{0.01\linewidth}%
\begin{subfigure}[t]{0.24\linewidth}\centering\includegraphics[width=\linewidth]{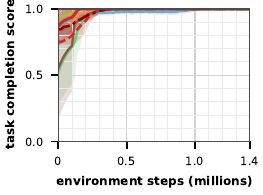}\caption{\rev{Robomimic-lift}}\end{subfigure}\\[2pt]
\begin{subfigure}[t]{0.24\linewidth}\centering\includegraphics[width=\linewidth]{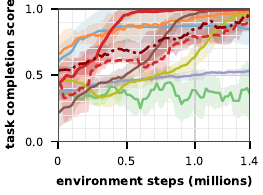}\caption{Kitchen-mixed}\end{subfigure}\hspace{0.01\linewidth}%
\begin{subfigure}[t]{0.24\linewidth}\centering\includegraphics[width=\linewidth]{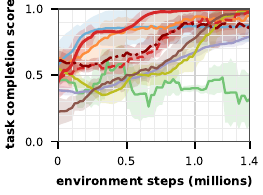}\caption{Kitchen-partial}\end{subfigure}\hspace{0.01\linewidth}%
\begin{subfigure}[t]{0.24\linewidth}\centering\includegraphics[width=\linewidth]{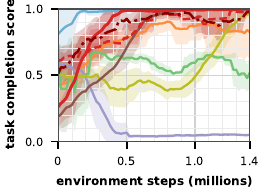}\caption{Kitchen-complete}\end{subfigure}\hspace{0.01\linewidth}%
\begin{subfigure}[t]{0.24\linewidth}\centering\includegraphics[width=\linewidth]{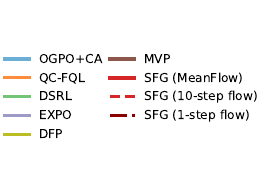}\end{subfigure}
\caption{Task completion score over the budget on the \rev{seven} tasks of Table~\ref{tab:main}, under the protocol of Section~\ref{sec:exp_setup}. \rev{SFG (MeanFlow) is the one-step version of SFG with $\delta = 0.01$, starting from the pretrained $\theta_0$ at step 0; SFG (10-step flow) starts from the same ten-step flow-matching policy as OGPO+CA, and SFG (1-step flow) from a one-step policy distilled from that ten-step policy.}}
\label{fig:main}
\vspace{6pt}
{\footnotesize\setlength{\tabcolsep}{6pt}%
\input{sections/table_main_v2_body}}
\captionof{table}{Main results, \rev{seven-task} means (per task: Table~\ref{tab:main_full}; \texttt{transport} is reported separately in Figure~\ref{fig:transport}): task completion score at the end of the budget and environment steps (M) to a score of 0.8 (1.4M if never), with whether each method freezes the pretrained policy and keeps an anchor (Appendix~\ref{app:baselines}).}
\label{tab:main}
\end{figure}
}

\paragraph{Baselines.}
We compare against six generative-policy fine-tuning methods: OGPO+CA~\citep{patil2026ogpo}, which fine-tunes the full policy; QC-FQL~\citep{li2025qc}, which distills a one-step actor; DSRL~\citep{wagenmaker2025dsrl} and EXPO~\citep{dong2025expo}, which steer or edit a frozen policy; and MVP~\citep{zhan2026mvp} and DFP~\citep{koo2026dfp}, which train one-step generators with anchors to the data.
Each baseline uses the authors' implementation and published defaults under the same interaction budget.
Table~\ref{tab:baselines} in Appendix~\ref{app:baselines} summarizes their update mechanisms and actor constraints.

\paragraph{Ours.}
Unless otherwise stated, we use the one-step version of SFG (Algorithm~\ref{alg:direct}): the MeanFlow generator pretrained to convergence, $\delta = 0.01$, and an actor learning rate of $3 \times 10^{-7}$ on every task. %
The multi-step versions (Appendix~\ref{app:ode}) appear on \texttt{transport} (Section~\ref{sec:exp_main}) and in the sampler ablation (Section~\ref{sec:exp_ablation}).
\rev{The main results (Figure~\ref{fig:main}, Table~\ref{tab:main}) evaluate three variants of SFG: this one-step MeanFlow version; SFG (10-step flow), which starts from the same ten-step flow-matching policy as OGPO+CA; and SFG (1-step flow), a one-step policy distilled from that ten-step flow policy.}

\paragraph{Protocol.}
Every method gets 1.4M online environment steps and starts from its own behavior cloning policy trained to convergence, whose success is the point at step zero.
Policies are evaluated every 20k steps on 64 episodes; %
Every reported number is the mean over four seeds, for the baselines as for SFG. %

\subsection{Main Results}
\label{sec:exp_main}

SFG matches or exceeds the best baseline in final task completion score on all \rev{seven} tasks and reaches a score of 0.8 earliest on average (Table~\ref{tab:main}, Figure~\ref{fig:main}).
Its sample-efficiency gains are largest on \texttt{square} and \texttt{tool\_hang}, while OGPO+CA reaches the threshold earlier on \texttt{kitchen-complete}\rev{; on \texttt{can} and \texttt{lift}, which several methods solve within the first evaluations, the differences are small}.
This compares complete recipes; the differences are not attributed to the update rule or the bound alone.
The curves show what the endpoints do not: on \texttt{tool\_hang}, the three baselines that start from a pretrained policy as good as ours (DSRL, QC-FQL, EXPO) first fall to zero, whereas the smoothed \rev{evaluation} success of SFG never drops below its starting point; OGPO+CA, DFP, and MVP start from much weaker pretrained policies and climb.
\rev{The starting policy does not explain the ranking: SFG (MeanFlow) starts at an initial score of 0.43, the same as OGPO+CA and below DSRL, EXPO, and QC-FQL (0.53--0.57), and still reaches the best final score and the earliest 0.8. In the comparison matched to OGPO+CA, where SFG (1-step flow) starts from a one-step policy distilled from the same ten-step flow policy, SFG keeps the advantage (0.94 and 0.47M steps against 0.91 and 0.56M). Fine-tuning the ten-step flow policy itself is weaker than both one-step versions (0.92 and 0.68M), as the critic's gradient still has to be carried through a multi-step sampler, but it ends at a completion score comparable to the best baselines (0.93 for MVP, 0.91 for OGPO+CA).}

\newpage  %
\begin{wrapfigure}[16]{r}{0.44\linewidth}  %
\vspace{-\baselineskip}
\centering
\includegraphics[width=\linewidth]{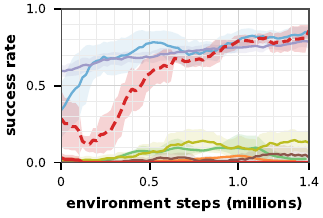}\\[-2pt]
\includegraphics[width=\linewidth]{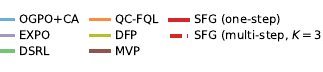}
\caption{Robomimic-\texttt{transport}: the one-step and the multi-step ($K = 3$) versions of SFG against the baselines.}
\label{fig:transport}
\vspace{-0.5\baselineskip}
\end{wrapfigure}
\textbf{Transport.}
On \texttt{transport} the evaluated one-step methods achieve low final success: the one-step version of SFG, whose generator starts at 5\%, stays at zero at every evaluation, and QC-FQL, DFP, and MVP end below 15\%, as does DSRL (Figure~\ref{fig:transport}); only OGPO+CA and EXPO achieve substantially higher success.
We therefore run the multi-step version of SFG from a pretrained flow-matching policy decoded with three Euler steps: with the same direct $Q$-gradient and functional step bound it climbs from 54 to 81\%, below OGPO+CA's 87\% (other configurations: Appendix~\ref{app:transport}).
The two versions use different pretrained policies, so this comparison demonstrates applicability to multi-step sampling without isolating the effect of sampler depth. %

\subsection{Ablations}  %
\label{sec:exp_ablation}

\begin{figure}[t]
\centering
{\footnotesize\setlength{\tabcolsep}{5pt}\input{sections/table_ablate_v2_body}}
\captionof{table}{Ablations of SFG on \rev{\texttt{tool\_hang}, \texttt{kitchen-partial}, and \texttt{square}}. \rev{\texttt{kitchen-partial} reports the fraction of the four subtasks completed, as in Table~\ref{tab:main}.}}
\label{tab:ablate}
\vspace{2pt}
\includegraphics[width=0.88\linewidth]{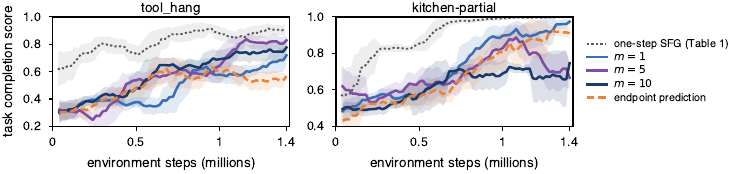}  %
\captionof{figure}{SFG on a ten-step flow policy, by the rule that carries the critic's gradient through the sampler: backpropagation through the last $m$ steps (Algorithm~\ref{alg:ode_bptt}) or endpoint prediction (Algorithm~\ref{alg:ode_jump}); the one-step reference is the SFG run of Table~\ref{tab:main}, a separately pretrained MeanFlow policy. \texttt{kitchen-partial} reports the mean fraction of subtasks completed, as in Table~\ref{tab:main}.}
\label{fig:multi-step}
\end{figure}

Table~\ref{tab:ablate} removes or varies one component of SFG at a time on \texttt{tool\_hang}, where the pretrained policy succeeds half the time, and \texttt{kitchen-partial}, where it never completes all four subtasks.
Cells are the success rate over the first 0.3M steps and over the last 200k steps as in Table~\ref{tab:main}, each the mean over four seeds, at the learning rate of $3 \times 10^{-7}$ except in the learning-rate rows. The comparison with and without the bound fixes $\tau = 0.05$; curves and details are in Appendix~\ref{app:fix}.

\textbf{The functional step bound.}
Without it, the direct update loses the pretrained competence on \texttt{tool\_hang}, staying at zero for 0.4M steps before recovering to the same endpoint; with $\delta = 0.01$ \rev{its smoothed evaluation success} never falls below $\theta_0$.
On \texttt{kitchen-partial}, where the pretrained policy has zero full-task success, the run without the bound improves earlier, while both variants reach similar final performance; the clearest benefit of the bound is therefore the preservation of early performance on \texttt{tool\_hang}.
Changing the bound alters the trade-off between early preservation and subsequent improvement; we use $\delta = 0.01$ as the default. \rev{Appendix~\ref{app:trust_empirical} compares the projection with a penalty on the action change and a parameter-space trust region at a matched step.}

\textbf{Learning rate.}
Among the tested learning rates, none without the bound preserves the early success of the run with the bound on \texttt{tool\_hang}, and the rates small enough to help there slow learning or lower the final performance on \texttt{kitchen-partial}: a fixed rate cannot adapt the step to the task and the stage of training (Section~\ref{sec:analysis_step}).

\textbf{Demonstration replay.}
Removing the demonstration half of every batch leaves \texttt{tool\_hang} at 0.65 and \texttt{kitchen-partial} at zero, and removing the bound as well gives zero on both: the bound limits how far an update moves the policy, and demonstration replay supplies successful experience for value learning while online successes are scarce.

\input{sections/experiments_bucket}

\textbf{Multi-step samplers.}
All variants execute the same ten-step sampler; the backpropagation depth $m$ determines how many of its final steps receive gradients, and endpoint prediction instead evaluates the critic on actions predicted from intermediate states, without backpropagating through the sampling chain.
SFG also improves ten-step flow policies with full or truncated backpropagation (Figure~\ref{fig:multi-step}).
Part of the \texttt{tool\_hang} runs were throttled by the projection solver (Appendix~\ref{app:ode}), so we do not rank the depths.
These runs show that the update trains through a multi-step sampler, not that it matches the stability or the final performance of the one-step version; Appendix~\ref{app:ode} reports all six tasks (Table~\ref{tab:multi-step}). %

%% file: sections/table_main_v2_body.tex
\begin{tabular}{@{}l cc ccc@{}}
\toprule
Method & Policy frozen & Anchor & \rev{Initial score} & Completion score & Steps to 0.8 (M) \\
\midrule
DSRL & yes & no & \rev{0.53} & \rev{0.59} & \rev{1.18} \\
EXPO & yes & yes & \rev{0.57} & \rev{0.63} & \rev{0.98} \\
QC-FQL & no & yes & \rev{0.54} & \rev{0.68} & \rev{0.77} \\
DFP & no & yes & \rev{0.41} & \rev{0.88} & \rev{0.76} \\
MVP & no & yes & \rev{0.28} & \rev{0.93} & \rev{0.64} \\
OGPO+CA & no & yes & \rev{0.43} & \rev{0.91} & \rev{0.56} \\
\rev{SFG (10-step flow)} & \rev{no} & \rev{no} & \rev{0.43} & \rev{0.92} & \rev{0.68} \\
\rev{SFG (1-step flow)} & \rev{no} & \rev{no} & \rev{0.41} & \rev{0.94} & \rev{0.47} \\
\rev{SFG (MeanFlow)} & no & \textbf{no} & \rev{0.43} & \rev{\textbf{0.98}} & \rev{\textbf{0.21}} \\
\bottomrule
\end{tabular}

%% file: sections/table_ablate_v2_body.tex
\begin{tabular}{@{}l l cc cc cc@{}}
\toprule
& & \multicolumn{2}{c}{\texttt{tool\_hang}} & \multicolumn{2}{c}{\texttt{kitchen-partial}} & \multicolumn{2}{c}{\rev{\texttt{square}}} \\
\cmidrule(lr){3-4}\cmidrule(lr){5-6}\cmidrule(lr){7-8}
& & first 0.3M & final & first 0.3M & final & \rev{first 0.3M} & \rev{final} \\
\midrule
pretrained $\theta_0$ & & 0.50 & & \rev{0.32} & & \rev{0.19} & \\
SFG & $\delta = 0.01$ & \rev{0.70} & \rev{0.91} & \rev{0.70} & \rev{1.00} & \rev{0.73} & \rev{0.98} \\
\midrule
bound & none & 0.00 & 0.90 & \rev{0.65} & \rev{0.99} & \rev{0.69} & \rev{0.99} \\
& $\delta = 0.005$ & 0.69 & 0.83 & \rev{0.57} & \rev{0.94} & \rev{0.71} & \rev{0.98} \\
& $\delta = 0.05$ & 0.37 & 0.83 & \rev{0.84} & \rev{0.98} & \rev{0.79} & \rev{0.98} \\
\midrule
no bound, & \rev{$3 \times 10^{-7}$} & \multicolumn{2}{c}{\rev{(= none)}} & \multicolumn{2}{c}{\rev{(= none)}} & \multicolumn{2}{c}{\rev{(= none)}} \\
learning rate & $10^{-7}$ & 0.22 & 0.84 & \rev{0.47} & \rev{0.99} & \rev{0.67} & \rev{0.97} \\
& $3 \times 10^{-8}$ & 0.56 & 0.88 & \rev{0.53} & \rev{0.91} & \rev{0.72} & \rev{0.97} \\
& $10^{-8}$ & 0.43 & 0.85 & \rev{0.53} & \rev{0.70} & \rev{0.68} & \rev{0.98} \\
\midrule
data & no demonstration replay & 0.48 & 0.65 & \rev{0.42} & \rev{0.50} & \rev{0.43} & \rev{0.98} \\
\bottomrule
\end{tabular}

%% file: sections/experiments_bucket.tex
\label{sec:exp_bucket}

\textbf{Beyond state inputs.}
\rev{SFG also fine-tunes policies that act on high-dimensional observations. On four dexterous manipulation tasks whose observations are point clouds computed from a single camera view, DexArt \texttt{bucket}~\citep{bao2023dexart} and Adroit \texttt{door}, \texttt{pen}, and \texttt{relocate}~\citep{rajeswaran2018dapg}, with the 255M-parameter architecture of DP3~\citep{ze2024dp3} and a MeanFlow head, SFG raises the success rate of the pretrained policy from 58\% to 78\% on average at the end of training, a gain of 20 points (Figure~\ref{fig:bucket}). The comparison with the PPO demonstrator on \texttt{bucket} and the run without the bound are in Appendix~\ref{app:fix}.}

\noindent\begin{minipage}{\linewidth}
\captionsetup{type=figure}
\centering
\begin{subfigure}[t]{0.24\linewidth}\centering\includegraphics[width=\linewidth]{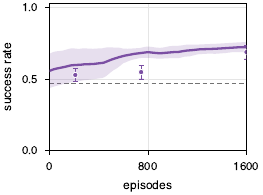}\caption{DexArt-\texttt{bucket}}\end{subfigure}\hspace{0.01\linewidth}%
\begin{subfigure}[t]{0.24\linewidth}\centering\includegraphics[width=\linewidth]{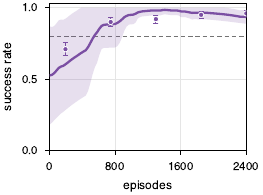}\caption{\rev{Adroit-\texttt{door}}}\end{subfigure}\hspace{0.01\linewidth}%
\begin{subfigure}[t]{0.24\linewidth}\centering\includegraphics[width=\linewidth]{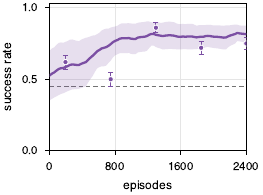}\caption{\rev{Adroit-\texttt{pen}}}\end{subfigure}\hspace{0.01\linewidth}%
\begin{subfigure}[t]{0.24\linewidth}\centering\includegraphics[width=\linewidth]{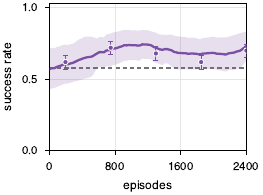}\caption{\rev{Adroit-\texttt{relocate}}}\end{subfigure}\\[1pt]
\includegraphics[width=0.92\linewidth]{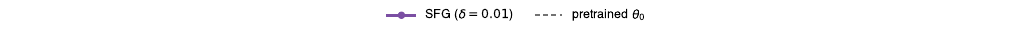}
\caption{DexArt-\texttt{bucket} \rev{and three Adroit tasks} from point clouds. \rev{Adroit: 100-episode holdouts, against $\theta_0$ evaluated on the same episodes.}}
\label{fig:bucket}
\end{minipage}

%% file: sections/related_work.tex
\section{Related Work}
\label{sec:related}

\paragraph{RL fine-tuning of generative policies.}
On-policy methods apply policy gradients to the denoising chain as an inner MDP~\citep{ren2024dppo,zhang2025reinflow} or through a flow-matching surrogate for the likelihood ratio~\citep{mcallister2025fpo}, at a high sample cost~\citep{patil2026ogpo}. %
Most off-policy methods avoid differentiating the critic through the sampler: they learn a correction around the fixed pretrained policy~\citep{wagenmaker2025dsrl,dong2025expo,li2025qc,tang2026awls}, extract a policy from the critic by reweighting, distillation, or a one-step actor~\citep{hansenestruch2023idql,mark2024parl,park2025fql}, or replace the critic gradient by a surrogate over the chain~\citep{patil2026ogpo,doo2026qflow,wang2026qvgm}; those that do differentiate through it keep a behavior cloning or proximity term~\citep{wang2023diffusionql,zhang2026sacflow}. One-step MeanFlow generators remove the chain but keep a term on the data, flow matching in OFQL, MeanFlowQL, and MVP~\citep{nguyen2026ofql,wang2026meanflowql,zhan2026mvp} and behavior cloning in DFP~\citep{koo2026dfp}. Our actor is the pretrained generator itself, updated by the critic gradient alone; Appendix~\ref{app:baselines} tabulates what each of these methods trains and what holds it. %

\paragraph{Why fine-tuning fails, and the anchor.}
Prior work attributes failures to backpropagation through the sampler~\citep{park2025fql,dong2025expo,zhang2026sacflow}, to critic errors that an expressive actor over-optimizes~\citep{patil2026ogpo,zhang2025d2ac,dong2026trqam}, to the offline-to-online transition~\citep{nakamoto2023calql,zhou2024wsrl,wolczyk2024forgetting}, to exploration under sparse reward~\citep{zhang2025reinflow,zhang2026sacflow}, and to saturation of bounded actions~\citep{hausknecht2016parameterized,wang2026meanflowql}. The common remedy pulls the policy toward the data: a behavior cloning loss inherited from offline RL~\citep{wang2023diffusionql,hansenestruch2023idql,park2025fql}, a proximity penalty~\citep{zhang2026sacflow}, or a loss on successful online episodes~\citep{patil2026ogpo}. %
For Gaussian or deterministic policies, TD3-C and PROTO replace this term by a bound on the step from the previous policy~\citep{luo2023td3c,li2023proto}. %
We study whether controlling changes in generated actions can likewise support fine-tuning of pretrained flow policies without an actor anchor. Existing constraints for pretrained flow policies penalize deviation from the pretrained policy~\citep{ziakas2026fmq,dong2026trqam,sun2026dicerl}, whereas trust regions centered at the previous iterate bound each step~\citep{schulman2015trpo,sun2025swfp}; SFG measures the change in generated actions relative to the previous committed actor, so the reference advances during fine-tuning (Appendix~\ref{app:trust}).

%% file: sections/conclusion.tex
\section{Conclusion and Limitations}
\label{sec:conclusion}

We asked whether the direct $Q$-gradient can fine-tune a pretrained flow policy without an anchor.
On a one-step generator, a single sufficiently large update along the critic's gradient erases the pretrained competence; bounding that change once per episode, with no anchor, no distillation, and no extra network, improves the early evaluations of the committed policy in the matched comparison on \texttt{tool\_hang}.
The resulting recipe, SFG, starts from behavior cloning policies trained to convergence, reaches 80\% success sooner and ends higher than the compared methods on \rev{seven} manipulation tasks, and carries over to multi-step flow policies and to point-cloud inputs.
In these settings, an actor anchor toward the demonstrations or the pretrained policy was not needed for effective learning; what the bound adds is the preservation of the pretrained competence through the first updates. %

Four limitations remain.
The ablations cover \rev{three} tasks and do not apportion the \rev{seven-task} gains among the components of the full configuration; the critic aggregation and the loss normalization are listed as settings (Appendix~\ref{app:impl}) rather than ablated.
All experiments are in simulation, with state observations except for \rev{four} point-cloud tasks; nothing in the recipe is specific to simulation, and a real-robot study, \rev{for instance on contact-rich manipulation~\citep{he2026frvla},} is the natural next step. %
The multi-step version of SFG trains, but its runs neither match the final performance of the one-step version nor establish the same stability, and part of the \texttt{tool\_hang} runs were affected by the projection solver (Appendix~\ref{app:ode}); they show applicability only.
Finally, the bound $\delta$ is a fixed value chosen once and reused across tasks, and choosing it took some tuning (Table~\ref{tab:ablate}); adapting it automatically during fine-tuning is a direction we plan to explore.

%% file: sections/appendix_protocol.tex
\section{Analysis Protocol}
\label{app:protocol}

\paragraph{Task.} Section~\ref{sec:analysis} uses \texttt{tool\_hang} (proficient-human demonstrations, low-dimensional observations) from Robomimic, chosen because the pretrained policy leaves ample room to improve.

\paragraph{Pretrained policy.} Every run starts from the same generator $\theta_0$, pretrained to convergence with the MeanFlow objective on the 200 demonstrations. On a fixed probe set of 64 episodes it succeeds in 41 to 47\% of episodes across five readings, and in 49\% on its own training-time evaluation. %
Online, the parameters $\theta$ of the action decoder are updated and the observation encoder is kept frozen.

\paragraph{Learning rate.} Every run of Section~\ref{sec:analysis} uses a constant actor learning rate of $4.5\times10^{-5}$ unless stated otherwise, below the $3\times10^{-4}$ default of TD3 and SAC~\citep{fujimoto2018td3,haarnoja2018sac}; the critic learning rate is $3\times10^{-4}$.

\paragraph{Critic and replay.} The critic is the ensemble of Section~\ref{sec:prelim_rlpd} with $N = 10$ heads and random subsets of $M = 2$; every batch of 256 transitions draws 128 from the demonstrations and 128 from the online buffer.

\paragraph{Anchor.} The anchored runs of Figure~\ref{fig:failure} add to the actor loss~\eqref{eq:actor} the behavior cloning term $\mathcal{R}(\theta;\mathcal{D}) = \E_{(s,a) \sim \mathcal{D}_{\mathrm{demo}},\,\epsilon}\,\|f_\theta(s,\epsilon) - a\|^2$, the squared error between the generated and the demonstrated action chunk, at weights $\lambda = 1$ and $3$; every other run sets $\lambda = 0$.

\paragraph{Probe bank.} The bank $\mathcal{B}$ of~\eqref{eq:fstep} holds 32 states drawn once from the demonstrations with a fixed seed, each paired with 8 noise draws, 256 rows in all. The norm in~\eqref{eq:fstep} is the Euclidean norm of the whole action chunk (7 action dimensions over a horizon of 8 on \texttt{tool\_hang}), and $d$ is the root mean square of these norms over the bank. The probe of Figure~\ref{fig:step}, the proposals of Figure~\ref{fig:raw}, and the functional step bound $\delta$ of Section~\ref{sec:method} are all measured in this one quantity.

\paragraph{Budget and evaluation.} Runs use one million environment steps unless stated otherwise. Two measurements are reported and never mixed. Training curves are evaluations of 64 episodes taken every 20{,}000 steps; where a single training-curve number is quoted it is the mean of the last eight evaluations at or below the stated budget (512 episodes). Final success rates are measured on the final checkpoint with 100 held-out episodes and fixed reset seeds. Evaluation labels count online steps plus the 10{,}000 critic warm-up steps. With the bound in place, each evaluation is taken at the first episode boundary at or after its scheduled step, on the committed actor. %

\paragraph{Seed-to-seed spread.} Four seeds of one configuration (the unanchored direct update at $4.5\times10^{-5}$) end at 33, 35, 48, and 55\% on the held-out evaluation. %

\paragraph{Probe.} The single-step probe of Figure~\ref{fig:step} takes the critic at the end of a successful run (its policy succeeds in 93\% of episodes), freezes it, computes the direct-update gradient once at $\theta_0$ on a batch of 256 demonstration observations, and evaluates $\theta_0 - L\,\hat g$ for a grid of $L$ on the same 64-episode probe set with the same reset seeds. The functional step on the horizontal axis is the same quantity that the projection of Section~\ref{sec:method} bounds. Five independent gradient draws were probed; Figure~\ref{fig:step} shows three of them singly and the mean of all five.

\paragraph{Figure~\ref{fig:failure_step}.} Panel (a) shows four seeds of the direct update with RLPD, two seeds at each anchor weight ($\lambda = 1, 3$), two without RLPD, four at the 150-fold smaller learning rate (target rate $\tau = 0.005$), and two at that learning rate with the functional step bound of Section~\ref{sec:method} ($\delta = 0.01$, target rate $\tau = 0.05$; this is a method reference, since $\tau$ differs; the comparison at matched $\tau$ is the bound row of Table~\ref{tab:ablate} and Appendix~\ref{app:fix}; the second run stops at 430k steps, and the shading ends with it); lines are the mean of the evaluations in a centered window of a tenth of the budget (five evaluations at the larger learning rate, seven at the smaller), pooled over seeds, shading is one standard deviation of the same pool, and the gray band spans the five readings of $\theta_0$ above. Panel (b) shows the five gradient draws of the probe, three of them singly and all five in the mean, with one standard deviation across draws. The dashed curve moves $\theta_0$ along three random unit directions in parameter space, each scaled so that its functional step on the probe bank matches the abscissa (within 1\%); a random direction needs about a hundred times the parameter displacement of the gradient for the same functional step. Panel (c) shows the functional step proposed in each episode as medians over 50{,}000-step bins, with lines and shading as in (a), for the runs of (a) without the bound at both learning rates; the gray interval is the phase in which the unanchored runs of (a) have zero success. %

\section{Drift in Action Space}
\label{app:drift}

\begin{figure}[H]
\centering
\includegraphics[width=0.9\linewidth]{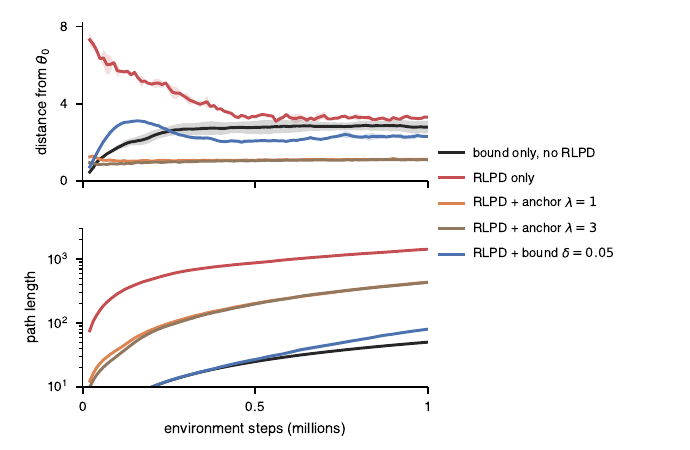}
\caption{Drift in action space, in the metric that the projection bounds, for the runs of Figure~\ref{fig:failure} at the learning rate of Section~\ref{sec:analysis} ($4.5\times10^{-5}$): the update without the bound, the anchor at $\lambda = 1$ and $3$, the bound at $\delta = 0.05$ (the value of those runs), and the bound without RLPD. Top: net distance of the policy from $\theta_0$. Bottom: cumulative path length, log scale. Lines are means, shading one standard deviation across seeds.}
\label{fig:drift_app}
\end{figure}

Figure~\ref{fig:drift_app} separates what the anchor and the bound control, at the learning rate and bound of Section~\ref{sec:analysis} ($4.5\times10^{-5}$, $\delta = 0.05$). The anchor keeps the policy near $\theta_0$ (a net distance of about 1.1 throughout), the bound does not (2.3, against 3.3 without it); what the bound changes is the path, whose length falls from about 1{,}400 without the bound to 80 with it, while the anchor lies in between at about 430. The update without the bound first moves far from $\theta_0$ (above 7 within the first episodes) and returns only slowly, which is the collapse and partial recovery of Figure~\ref{fig:failure} seen from the policy's side.

\paragraph{At the learning rate of the main results.} At $3\times10^{-7}$ with $\delta = 0.01$ the projection binds at every episode boundary, so the committed step equals the bound and the drift is set by the number of episodes and the direction of the steps. Adding the anchor of Section~\ref{sec:analysis_failure} on top of the bound changes what the optimizer proposes (Table~\ref{tab:drift_main}): over the first 300k steps, the raw functional step per episode falls from 0.15 without the anchor to 0.08 at $\lambda = 3$ and 0.018 at $\lambda = 150$ on \texttt{tool\_hang}, and on \texttt{kitchen-partial} from 0.064 to 0.008 at $\lambda = 150$, below the bound itself, so that the projection binds in 11\% of episodes (2\% after 900k steps). The net distance from $\theta_0$ after 300k steps is 1.00 without the anchor against 0.61 at $\lambda = 3$ and 0.73 at $\lambda = 150$ on \texttt{tool\_hang}, and 0.60 against 0.36 on \texttt{kitchen-partial}, where the anchored policy then stays at 0.38 while the unanchored one moves out to 1.78 by the end of the budget.

\begin{table}[H]
\centering
\small
\begin{tabular}{@{}l c c c c c@{}}
\toprule
& $\lambda$ & episodes & raw step & distance from $\theta_0$ & projected \\
\midrule
\texttt{tool\_hang}, first 300k & 0 & 445 & 0.152 & 1.00 & 100\% \\
& 3 & 457 & 0.077 & 0.61 & 100\% \\
& 150 & 322 & 0.018 & 0.73 & 99\% \\
\midrule
\texttt{kitchen-partial}, first 300k & 0 & 524 & 0.064 & 0.60 & 100\% \\
& 150 & 501 & 0.008 & 0.36 & 11\% \\
\texttt{kitchen-partial}, after 900k & 0 & 1732 & 0.023 & 1.78 & 100\% \\
& 150 & 901 & 0.007 & 0.38 & 2\% \\
\bottomrule
\end{tabular}
\caption{The bound with and without the anchor at the main line ($3\times10^{-7}$, $\delta = 0.01$, $\tau = 0.05$): episodes in the window, median raw functional step proposed per episode, net distance from $\theta_0$ in action space at the end of the window, and the fraction of episodes in which the projection binds. The first-300k rows cover the same 300k environment steps, over which the arms complete different numbers of episodes; the last two rows cover the run from 900k steps on.}
\label{tab:drift_main}
\end{table}

\section{Training Curves for the Ablations}
\label{app:fix}

\begin{figure}[H]
\centering
\includegraphics[width=0.8\linewidth]{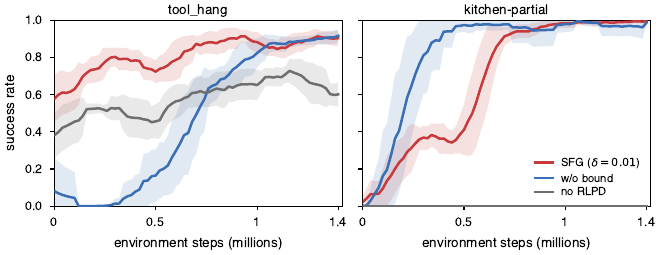}
\caption{Curves of the bound and demonstration arms of Table~\ref{tab:ablate}, smoothing and shading as in Figure~\ref{fig:main}; on \texttt{kitchen-partial}, success is all four subtasks completed, as in that table.}
\label{fig:ablate}
\end{figure}

\begin{figure}[H]
\centering
\includegraphics[width=\linewidth]{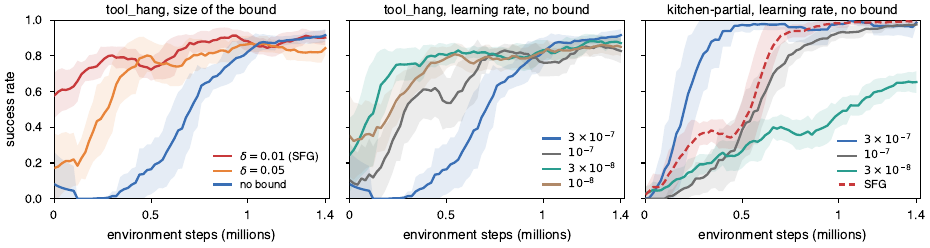}
\caption{The bound-size and learning-rate rows of Table~\ref{tab:ablate}: (a) \texttt{tool\_hang} with $\delta = 0.01$, $\delta = 0.05$, and no bound; (b) \texttt{tool\_hang} and (c) \texttt{kitchen-partial} at smaller learning rates without the bound, with the SFG run of (c) dashed for reference. Smoothing and shading as in Figure~\ref{fig:main}. The $\delta = 0.005$ rows, the \texttt{kitchen-partial} $\delta = 0.05$ row, and the $10^{-8}$ row on \texttt{kitchen-partial} are reported in the table only.}
\label{fig:ablate_more}
\end{figure}

Figure~\ref{fig:ablate} shows the runs behind the bound and demonstration rows of Table~\ref{tab:ablate}, and Figure~\ref{fig:ablate_more} the bound-size and learning-rate rows, at the learning rates of the main results. %

\paragraph{Details of Table~\ref{tab:ablate}.} Without the bound, success on \texttt{tool\_hang} is zero over the first 0.3M steps and the first successful evaluation comes at 400k steps. On \texttt{kitchen-partial} the run with the bound climbs later than the one without (0.21 against 0.36 over the first 0.3M steps) and ends two points higher. A tighter bound ($\delta = 0.005$) keeps more of the early success on \texttt{tool\_hang} (0.69) but ends lower on both tasks (0.83 and 0.85); a looser one ($\delta = 0.05$) is the fastest on \texttt{kitchen-partial} (0.66 early) but already loses a third of the pretrained success on \texttt{tool\_hang} (0.37). Among the smaller learning rates without the bound, the best early success on \texttt{tool\_hang} is 0.56 at $3 \times 10^{-8}$, against 0.65 with the bound, and on \texttt{kitchen-partial} training stalls at 0.66 with $3 \times 10^{-8}$ and does not start at $10^{-8}$. Removing the demonstration half of every batch with the bound in place leaves \texttt{tool\_hang} at 0.65 and \texttt{kitchen-partial} at zero throughout; removing RLPD and the bound together gives zero on both tasks. A further variant, not in Table~\ref{tab:ablate}, adds the behavior cloning term of Section~\ref{sec:analysis_failure} to the update with the bound at $\lambda = 150$, the weight of that section raised by the ratio of the two learning rates; it ends at 0.20 and 0.24 on \texttt{tool\_hang} and 0.01 and 0.13 on \texttt{kitchen-partial}, against 0.65 and 0.90 and 0.21 and 0.99 with the bound alone.

\begin{figure}[H]
\centering
\includegraphics[width=0.48\linewidth]{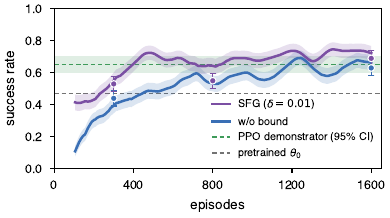}  %
\caption{\rev{DexArt-\texttt{bucket} from point clouds, with and without the bound: run-log curves (smoothed over 105 episodes), 100-episode holdouts as dots, and the PPO demonstrator's 95\% interval as the green band.}}
\label{fig:bucket_holdouts}
\end{figure}

\paragraph{\rev{DexArt \texttt{bucket}.}} \rev{The pretrained policy succeeds in 47\% of episodes and its PPO demonstrator in 65\% (Figure~\ref{fig:bucket_holdouts}; dots are 100-episode holdouts, the band the demonstrator's 95\% interval). With the bound, success never collapses and reaches the demonstrator's level within about 400 episodes; without it, success first falls close to zero and regains the pretrained level only after about 450 episodes, even at a hundredfold smaller learning rate.}

\section{The Remaining Runs on \texttt{transport}}
\label{app:transport}

\begin{figure}[H]
\centering
\includegraphics[width=0.32\linewidth]{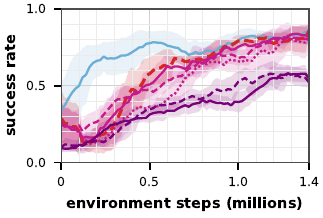}\hspace{0.03\linewidth}\includegraphics[width=0.32\linewidth]{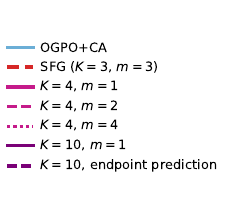}
\caption{The other multi-step configurations of SFG on \texttt{transport}: the flow-matching policy of OGPO decoded with $K = 4$ Euler steps and trained through the last one, two, or all four of them (Algorithm~\ref{alg:ode_bptt}), and decoded with $K = 10$ steps and trained through the last step or by the endpoint estimator (Algorithm~\ref{alg:ode_jump}); the $K = 3$ run of Figure~\ref{fig:transport} and OGPO+CA for reference. Four seeds each, $\delta = 0.01$, conventions as in Figure~\ref{fig:main} with $\theta_0$ at step 0.}
\label{fig:transport_variants}
\end{figure}

Figure~\ref{fig:transport} shows the two versions of SFG and every baseline on \texttt{transport}; the remaining multi-step configurations are collected here. Among the multi-step configurations (Figure~\ref{fig:transport_variants}), the runs with three or four Euler steps share one shape, a drop from $\theta_0$ to about 10\% in the first 200k steps, a return to 50\% by 500k, and a slow climb to about 80\% at 1.4M; how many of the steps the gradient passes through changes the endpoint by less than the spread of a single 64-episode evaluation (81 to 83\%). The ten-step policy starts at 13\% and reaches 57\% with the gradient through the last step and with the endpoint estimator alike. The $K = 3$ policy without the bound is at zero from its first evaluation on (not drawn). Of the baselines in Figure~\ref{fig:transport}, DFP and MVP, the two one-step generators anchored to the data, end at 13 and 5\%, and DSRL and QC-FQL stay below 5\% throughout, as does the one-step version of SFG, whose run was stopped at 950k steps after nineteen zero evaluations.

%% file: sections/appendix_meanflow.tex
\section{MeanFlow Pretraining Details}
\label{app:meanflow}

The average velocity of the flow of Section~\ref{sec:prelim_meanflow} over an interval is $u(z_t,r,t) = \frac{1}{t-r}\int_r^t v(z_\tau,\tau)\,\mathrm{d}\tau$, so that $z_r = z_t - (t-r)\,u(z_t,r,t)$ holds exactly for every $r < t$. Differentiating this definition in $t$ gives the MeanFlow identity
\begin{equation}
u(z_t,r,t) = v(z_t,t) - (t-r)\,\frac{\mathrm{d}}{\mathrm{d}t}\,u(z_t,r,t), \qquad \frac{\mathrm{d}}{\mathrm{d}t}\,u = v\,\partial_{z}u + \partial_t u,
\label{eq:mf_identity}
\end{equation}
where the total derivative is taken along the flow with $r$ held fixed. The network $u_\theta(z_t,r,t \mid s)$ is trained to satisfy this identity by regressing onto its right-hand side, with the conditional velocity $\epsilon - a$ in place of $v$ and the total derivative evaluated by a Jacobian-vector product whose gradient is stopped,
\begin{equation}
\mathcal{L}_{\mathrm{MF}}(\theta) = \E_{(s,a),\,\epsilon,\,(r,t)} \Big\| u_\theta(z_t,r,t \mid s) - \mathrm{sg}\Big[ (\epsilon - a) - (t-r)\,\frac{\mathrm{d}}{\mathrm{d}t}\,u_\theta(z_t,r,t \mid s) \Big] \Big\|^2 ,
\label{eq:mf_loss}
\end{equation}
with $r \le t$; at $r = t$ the target reduces to the flow matching target of~\eqref{eq:fm}.
We use this objective in the form of \citet{geng2025meanflow}, with three additions that we state for completeness.
\emph{Time sampling.} The pair $(r,t)$ is drawn with $r = t$ on half of the samples and $r < t$ on the other half.
\emph{Guidance.} The conditional velocity $\epsilon - a$ in the target is replaced by the guided velocity $\tilde v = w\,(\epsilon - a) + (1 - w)\, u_\theta(z_t,t,t \mid s)$ with $w = 2$, which folds classifier-free guidance into training so that inference still needs a single evaluation.
\emph{Loss weighting.} The squared error is weighted adaptively by $1 / (\|\cdot\|^2 + c)^{p}$ with $p = 0.5$ and $c = 10^{-3}$, computed without gradient.
The latent is $\epsilon \sim \mathcal{N}(0,I)$ of the shape of an action chunk. Observations and actions are normalized to $[-1,1]$ per dimension. Training uses Adam with learning rate $10^{-4}$, no weight decay, batch size $128$, gradient-norm clipping at $1.0$, and an exponential moving average of the weights with decay $0.999$; the averaged weights are the policy that is evaluated, saved, and fine-tuned. Checkpoints are screened on rollouts, never on the training loss, which does not track success (Appendix~\ref{app:protocol}). The decoder is an MLP of five layers of 512 units on \texttt{tool\_hang} and four on \texttt{square} and the kitchen tasks (Table~\ref{tab:hparams}). The checkpoints fine-tuned in Section~\ref{sec:experiments} are taken after 200{,}000 gradient steps on every task. %

%% file: sections/appendix_impl.tex
\section{Implementation Details of SFG}
\label{app:impl}

\paragraph{Actor loss.} The actor minimizes the negative subsampled minimum of~\eqref{eq:actor}, divided by the batch mean of $|Q|$ with the gradient stopped, so that the scale of the returns does not enter the effective learning rate. Its states are those of the same mixed batch as the critic's (Section~\ref{sec:prelim_rlpd}), so demonstration states enter the loss while demonstration actions do not. The generator output is clipped to $\mathcal{A} = [-1,1]^{d}$ once, inside the decode, and everything downstream uses the clipped action: execution, the critic's training and evaluation, the target action of~\eqref{eq:td}, and the probe bank on which the functional step~\eqref{eq:fstep} is measured. The actor loss uses the exact gradient of the clip, so that a saturated coordinate receives no gradient; a straight-through estimator through the clip was worse. %
The observation encoder of the pretrained generator is frozen; only the action decoder is updated.

\paragraph{Critic.} The critic is the ensemble of Section~\ref{sec:prelim_rlpd} with $N = 10$ layer-normalized heads. For every update a subset of $M = 2$ heads is drawn without replacement for the target of~\eqref{eq:td} and, independently, another for the actor loss. The action in the target is generated by the current (proposal) actor with fresh noise. Target networks are updated by Polyak averaging with rate $\tau = 0.05$ ($0.005$ in the runs of Section~\ref{sec:analysis}); the critic learning rate is $3\times10^{-4}$, and every batch of 256 transitions draws 128 from the demonstrations and 128 from the online buffer. %

\paragraph{Warm-up.} Before the first actor update, $W = 10{,}000$ environment steps are collected with $\theta_0$ (the last incomplete episode is discarded), and the critic is then trained alone for $W$ updates on the same mixed batches as in training; evaluation labels count these steps. %

\paragraph{Chunks and updates.} The policy emits a chunk of $H$ actions every $H$ environment steps and executes it open loop. Transitions are stored per environment step and enter the online buffer when the episode ends; a batch samples $H$-step segments and forms the chunk transition $(s_t, a_t, r^{H}_t, s_{t+H})$ of~\eqref{eq:td}. The critic and the actor are updated once per environment step, after that step's action has been executed. %

\paragraph{Projection.} The functional step is measured on the probe bank of Appendix~\ref{app:protocol}. The episode is collected by the proposal actor $\tilde\theta$; at the episode boundary, if $d(\theta, \tilde\theta) > \delta$, the fraction $\alpha$ is found by a proportional scalar iteration, $\alpha \leftarrow \alpha\,\delta / d(\theta, \theta + \alpha(\tilde\theta - \theta))$ from $\alpha = \delta / d(\theta, \tilde\theta)$, limited to 1 and accepted once the committed step is within 5\% of $\delta$ on either side, so that a committed step may exceed $\delta$ by up to 5\%; if eight iterations do not reach the tolerance, the best iterate is committed (96\% of projected episodes converge after one evaluation and 4\% after two, and the fallback fired once in 123{,}760 episodes) %
We use an actor learning rate of $3\times10^{-7}$ and $\delta = 0.01$ on every task.

\paragraph{Hyperparameters.} Table~\ref{tab:hparams} lists the settings of every run of Section~\ref{sec:experiments}; the runs of Section~\ref{sec:analysis} differ only where Appendix~\ref{app:protocol} says so.

\begin{table}[H]
\centering
\small
\begin{tabular}{@{}l >{\raggedright\arraybackslash}p{10.2cm}@{}}
\toprule
Setting & Value \\
\midrule
Actor learning rate & $3\times10^{-7}$, Adam, constant; gradient norm clipped at 1 \\
Functional step bound $\delta$ & 0.01 on every task \\
Probe bank & 32 demonstration states $\times$ 8 noise draws, fixed seed \\
Projection & proportional iteration, tolerance $\pm5\%$ of $\delta$, at most 8 iterations; optimizer state kept \\
Actor loss & subsampled minimum over $M = 2$ critic heads, divided by the batch mean of $|Q|$ (gradient stopped) \\ %
Critic & $N = 10$ heads with layer normalization, $M = 2$ for the target and for the actor loss \\
Critic learning rate & $3\times10^{-4}$, Adam, weight decay $10^{-5}$; target rate $\tau = 0.05$ \\
Discount $\gamma$ & 0.99 (\texttt{square}, kitchen); 0.999 (\texttt{tool\_hang}, \texttt{transport}) \\
Chunk length $H$ & 4 (\texttt{square}, kitchen); 8 (\texttt{tool\_hang}, \texttt{transport}) \\
Networks & MLPs of $4 \times 512$ (\texttt{square}, kitchen) or $5 \times 512$ (\texttt{tool\_hang}, \texttt{transport}), actor and critic alike \\
Batch & 256 transitions, 128 from the demonstrations and 128 from the online buffer; actor batch 128 \\
Replay buffer & $2\times10^{6}$ transitions; demonstrations kept throughout \\
Warm-up & $W = 10{,}000$ steps with $\theta_0$, then $W$ critic updates \\
Updates & one critic and one actor update per environment step \\
Budget and evaluation & $1.4$M online steps; every 20k steps (10k on \texttt{square}), 64 episodes \\
Generator & one-step MeanFlow decode; observation encoder frozen \\
Multi-step version & flow-matching policy with $K = 3$ (\texttt{transport}) or 10 Euler steps, 4 updates per step, 50k warm-up, evaluation every 40k steps \\ %
\bottomrule
\end{tabular}
\caption{Hyperparameters of SFG. Values shared across tasks are stated once; the per-task values follow the pretraining configuration of each benchmark.}
\label{tab:hparams}
\end{table}

%% file: sections/appendix_ode.tex
\section{SFG with a Multi-Step Sampler}
\label{app:ode}

When the pretrained policy is a flow-matching velocity field $v_\theta$ (Section~\ref{sec:prelim_meanflow}) rather than a MeanFlow generator, an action is the endpoint of $K$ Euler steps from $z_1 = \epsilon$ to $t = 0$, and $f_\theta$ is a composition of $K$ evaluations of $v_\theta$.
Algorithm~\ref{alg:direct} is unchanged: the decode $f_{\tilde\theta}$ at lines 5 and 8, the actor loss at line 9, and the probe bank behind $d(\cdot,\cdot)$ all call the same $K$-step sampler. %
What is left open is how the critic's gradient is taken through the chain at line 9.
We use two rules, given as Algorithms~\ref{alg:ode_bptt} and~\ref{alg:ode_jump}.
Both leave the forward pass untouched: execution, the TD target, the probe bank, and evaluation all use the full $K$-step endpoint, so runs that differ only in the gradient rule execute the same policy at the same parameters and differ only in the direction of the update.
Here $\mathrm{sg}$ is the stop-gradient (the identity in the forward pass) and $\mathrm{clip}$ the action clipping to $\mathcal{A} = [-1,1]^{d}$ of Section~\ref{sec:prelim_meanflow}; only the final action is clipped, not the intermediate states.

\paragraph{Truncated backpropagation.}
Algorithm~\ref{alg:ode_bptt} backpropagates through the last $m$ of the $K$ steps and detaches the state entering them.
Since the velocity evaluated at a detached step reaches the action only through that state, the first $K - m$ evaluations of $v_\theta$ receive no gradient at all; with $m = K$ the gradient passes through the whole decoder, with $m = 1$ through its last evaluation only.
The backward pass stores the activations of $m$ chained network evaluations.

\paragraph{Endpoint prediction.}
Algorithm~\ref{alg:ode_jump} takes no gradient through the chain.
From each intermediate state $z$ at time $t_k$, $k \ge k_0$, it predicts the endpoint in a single Euler step of length $t_k$, $\hat a_k = \mathrm{clip}(z - t_k\, v_\theta(z, t_k \mid s))$ with $z$ detached, and the actor loss is the mean over these predictions of the normalized critic value, each differentiable through one network evaluation; this is the action approximation of EDP~\citep{kang2023edp} applied to a flow sampler.
The executed action is still the $K$-step endpoint; the predictions exist only to carry the gradient, and the critic is evaluated on all of them in one call on a $(K - k_0)$-fold batch.
With $k_0 = K - 1$ there is a single prediction, $\hat a_{K-1} = \mathrm{clip}(z - \Delta\, v_\theta)$, which is the executed action itself, and the rule coincides with Algorithm~\ref{alg:ode_bptt} at $m = 1$; the two rules differ only in the predictions taken from earlier steps.
The normalization is per prediction, because the step of length $t_k$ scales the gradient of the earliest prediction well above that of the last one (a factor of about five between $k = K/2$ and $k = K - 1$ at $K = 10$), so that an unnormalized mean would be dominated by the noisiest prediction. %

\begin{algorithm}[H]
\small
\caption{$K$-step decode with truncated backpropagation (replaces $f_{\tilde\theta}$ in Algorithm~\ref{alg:direct})}
\label{alg:ode_bptt}
\begin{algorithmic}[1]
\Require velocity field $v_\theta$; state $s$; noise $\epsilon$; sampler steps $K$; backpropagated steps $m \in \{1, \dots, K\}$
\State $z \gets \epsilon$; \quad $\Delta \gets 1/K$
\For{$k = 0, \dots, K-1$}
    \State $t_k \gets 1 - k/K$
    \State $z \gets z - \Delta\, v_\theta(z, t_k \mid s)$ \Comment{one Euler step toward $t = 0$}
    \If{$k < K - m$}
        \State $z \gets \mathrm{sg}(z)$ \Comment{the gradient stops here; the value of $z$ is unchanged}
    \EndIf
\EndFor
\State \Return $a \gets \mathrm{clip}(z)$ \Comment{executed, scored by the critic, and measured on the probe bank}
\end{algorithmic}
\end{algorithm}

\begin{algorithm}[H]
\small
\caption{Endpoint-prediction gradient (replaces line 9 of Algorithm~\ref{alg:direct})}
\label{alg:ode_jump}
\begin{algorithmic}[1]
\Require velocity field $v_\theta$; state $s$; noise $\epsilon$; sampler steps $K$; first predicted step $k_0$ (default $K/2$)
\State $z \gets \epsilon$; \quad $\Delta \gets 1/K$; \quad $\mathcal{P} \gets \emptyset$
\For{$k = 0, \dots, K-1$}
    \State $t_k \gets 1 - k/K$; \quad $v \gets v_\theta\big(\mathrm{sg}(z), t_k \mid s\big)$ \Comment{differentiable in $\theta$, not through the chain}
    \If{$k \ge k_0$}
        \State $\hat a_k \gets \mathrm{clip}\big(\mathrm{sg}(z) - t_k\, v\big)$; \quad add $\hat a_k$ to $\mathcal{P}$ \Comment{one step of length $t_k$ predicts the endpoint}
    \EndIf
    \State $z \gets \mathrm{sg}(z) - \Delta\, \mathrm{sg}(v)$ \Comment{the chain is advanced without gradient}
\EndFor
\State $a \gets \mathrm{clip}(z)$ \Comment{executed action, TD target, probe bank: the same value as Algorithm~\ref{alg:ode_bptt}}
\State update $\tilde\theta$ by one optimizer step on $-\frac{1}{|\mathcal{P}|} \sum_{\hat a_k \in \mathcal{P}} \E_{s,\epsilon}\big[\min_{i \in \mathcal{M}'} Q_{\phi_i}(s, \hat a_k)\big] \big/ \mathrm{sg}\,\E\,|Q(s, \hat a_k)|$
\end{algorithmic}
\end{algorithm}

\paragraph{Setup.}
The pretrained policies are the flow-matching policies of the OGPO codebase, trained by behavior cloning and decoded with $K = 10$ Euler steps ($K \in \{3, 4\}$ in Appendix~\ref{app:transport}); on \texttt{tool\_hang} this policy succeeds in 25\% of episodes, on \texttt{transport} in 13\%.
Runs follow the recipe of Appendix~\ref{app:impl}, with these settings: $\delta = 0.01$ and $\tau = 0.05$, an actor learning rate of $3\times10^{-7}$ on every task, %
a warm-up of 50{,}000 steps followed by 400{,}000 critic updates, four critic and four actor updates per environment step, and 1.4M online steps; evaluations of 64 episodes are taken every 40{,}000 steps (35 per run).
Every cell of Table~\ref{tab:multi-step} is the mean over four seeds. %

\paragraph{Projection.}
The proportional iteration of Appendix~\ref{app:impl} assumes that the functional step is close to linear in $\alpha$; through $K$ applications of the actor it is not, and at $K = 10$ the iteration exhausts its budget at every episode boundary and would commit about a third of $\delta$.
The runs of this appendix therefore fall back to a bisection in $\log\alpha$ on $[10^{-12}, 1]$ with 24 iterations whenever the iteration misses its tolerance; the solutions lie near $\alpha \approx 10^{-7}$, where a bisection in $\alpha$ itself would spend its whole budget reaching that magnitude.
The projection binds at every episode boundary in every run. The fallback is never reached on the kitchen runs, whose committed step is within 1\% of $\delta$ at the median (residual 0.011 to 0.013), and it is reached in 30 to 63\% of the episodes on \texttt{tool\_hang}; on the last-step arm ($m = 1$) there, the solver misses its tolerance in half of the episodes and, after 1M steps, commits a step of exactly zero in one episode of five (7\% of all episodes), so that run is the one whose update was throttled by the solver rather than by $\delta$. %
The $K = 3$ and $K = 4$ runs of Appendix~\ref{app:transport} commit $0.0101$ at the median as well, with the fallback in at most 5\% of episodes. %
Without the functional step bound, the $K$-step policy on \texttt{transport} is at zero from its first evaluation on (Appendix~\ref{app:transport}).

\begin{table}[H]
\centering
\footnotesize
\setlength{\tabcolsep}{5pt}
\input{sections/table_ode_body}
\caption{SFG on a ten-step flow-matching policy, by the rule that carries the critic's gradient through the sampler: backpropagation through the last $m$ of the ten Euler steps (Algorithm~\ref{alg:ode_bptt}) or endpoint prediction from the last five steps, which takes no gradient through the chain (Algorithm~\ref{alg:ode_jump}). Task completion score at the end of the budget under the convention of Table~\ref{tab:main}: mean of the evaluations at 1.2M steps and later, kitchen as the fraction of subtasks completed; four seeds per cell. The reference column is the SFG row of Table~\ref{tab:main} ($^{\dagger}$\texttt{transport}: the three-step policy of Figure~\ref{fig:transport}). The same runs on the all-four-subtasks metric are quoted in the text; no ordering is read from the \texttt{tool\_hang} row, two of whose runs hit the solver fallback.}
\label{tab:multi-step}
\end{table}

\paragraph{Results.}
On \texttt{tool\_hang} the three depths end at 66, 82, and 76\% for $m = 1$, $5$, and $10$; the last-step and whole-chain runs are the two whose committed steps the projection solver throttled (previous paragraph), so we read no ordering of $m$ from this task.
Endpoint prediction ends at 53\% on \texttt{tool\_hang}, at 0.92 of the subtasks on \texttt{kitchen-partial} (0.93 for the last-step rule; 74\% against 73\% on all four), and at 57\% on \texttt{transport}, the same as the last-step rule there.
The last-step rule at $K = 10$ also reaches 99\% on \texttt{square}, 0.98 of the subtasks on \texttt{kitchen-complete}, and 0.87 on \texttt{kitchen-mixed} (96\% and 55\% of episodes completing all four).
On \texttt{transport} the last-step rule reaches 57\% at $K = 10$ against 83\% at $K = 4$ (Appendix~\ref{app:transport}); the pretrained policy itself succeeds in 52 to 54\% of episodes when decoded with two or three steps and in 13\% at ten.

\paragraph{Details behind Section~\ref{sec:exp_ablation}.} Decoding the MeanFlow network of the main results with ten steps and backpropagating through all of them, the update on \texttt{tool\_hang} still trains without the bound: it leaves zero later than the one-step run, at about 600k steps against 350k, and ends at 78\% against 91\%, a gap inside the seed-to-seed spread of Appendix~\ref{app:protocol} (four seeds each). On the flow-matching policy decoded with ten steps (Figure~\ref{fig:multi-step}, Table~\ref{tab:multi-step}), backpropagating through all ten steps reaches 76\% on \texttt{tool\_hang} and 68\% of the subtasks on \texttt{kitchen-partial}, and no depth reaches the one-step generator on either task. With the gradient through the last step, the ten-step policy matches the one-step generator on \texttt{square} and \texttt{kitchen-complete} and falls short where its pretrained policy is weaker, succeeding in 25\% of episodes on \texttt{tool\_hang} and 13\% on \texttt{transport} against 52 and 54\% for the policies of Table~\ref{tab:main}.

%% file: sections/table_ode_body.tex
\begin{tabular}{@{}l c cccc@{}}
\toprule
& SFG, Table~\ref{tab:main} & \multicolumn{4}{c}{ten-step flow-matching policy, by gradient rule} \\
\cmidrule(lr){3-6}
& (one-step) & $m{=}1$ & $m{=}5$ & $m{=}10$ & endpoint prediction \\
\midrule
\texttt{tool\_hang} & 0.91 & 0.66 & 0.82 & 0.76 & 0.53 \\
\texttt{kitchen-partial} & 1.00 & \textbf{0.93} & 0.70 & 0.68 & 0.92 \\
\texttt{transport} & 0.81$^{\dagger}$ & \textbf{0.57} & -- & -- & \textbf{0.57} \\
\texttt{square} & 0.98 & 0.99 & -- & -- & -- \\
\texttt{kitchen-mixed} & 1.00 & 0.87 & -- & -- & -- \\
\texttt{kitchen-complete} & 1.00 & 0.98 & -- & -- & -- \\
\bottomrule
\end{tabular}

%% file: sections/appendix_baselines.tex
\section{Baselines and Related Methods: What Is Trained and What Holds the Policy}
\label{app:baselines}

Table~\ref{tab:baselines} places the baselines of Section~\ref{sec:exp_setup} on the two axes of this paper: whether the policy's own weights are trained, and what holds the policy near where it started. Table~\ref{tab:related_axes} does the same for the other methods of Section~\ref{sec:related}, adding how the critic's signal reaches the policy. %

\begin{table}[H]
\centering
\small
\begin{tabular}{@{}l l >{\raggedright\arraybackslash}p{5.9cm}@{}}
\toprule
Method & Trains the policy's weights & Anchor \\
\midrule
DSRL & no (frozen); a policy over its noise & none; the policy itself stays frozen \\
EXPO & no (frozen); a residual edit policy & the frozen policy, bounded edits \\
QC-FQL & a one-step actor beside the flow policy & distillation to the flow policy fit to the data \\
MVP & yes, one-step generator & flow-matching term on the data \\
DFP & yes, one-step generator & behavior cloning term on the data \\
OGPO+CA & yes, full policy through the chain & PPO clip to the previous policy; imitation of its own successful episodes \\
SFG (ours) & yes, full decoder & none; a bound on the functional step of each episode \\
\bottomrule
\end{tabular}
\caption{What each method trains and what keeps its policy near where it started. SFG uses no actor anchor toward the demonstrations or the pretrained policy; it controls the change relative to the previous committed actor through projection.}
\label{tab:baselines}
\end{table}

\begin{table}[!tb]
\centering
\footnotesize
\setlength{\tabcolsep}{3.5pt}
\input{sections/table_main_full_v2_body}  %
\caption{Per-task values behind Table~\ref{tab:main}: task completion score at the end of the 1.4M-step budget (success rate on Robomimic, fraction of the four subtasks completed on kitchen), and the number of environment steps at which the smoothed score first reaches 0.8 (``--'': not within the budget, counted as 1.4M in the mean). Each cell is the mean over four seeds. Best value per column in bold.}
\label{tab:main_full}
\end{table}
\begin{table}[!tb]
\centering
\footnotesize
\setlength{\tabcolsep}{3pt}
\begin{tabular}{@{}>{\raggedright\arraybackslash}p{1.9cm} >{\raggedright\arraybackslash}p{2.3cm} >{\raggedright\arraybackslash}p{4.3cm} >{\raggedright\arraybackslash}p{4.3cm}@{}}
\toprule
Method & Trains the policy's weights & Critic reaches the policy by & Anchor or constraint (reference) \\
\midrule
Diffusion-QL & yes, through the chain & direct $Q$-gradient through the chain & denoising BC loss on the data (data) \\
IDQL & BC fit only & reweighting its samples & the policy is the BC fit (data) \\
FQL & a one-step actor beside the BC flow & direct $Q$-gradient into the one-step actor & distillation to the BC flow (data) \\
PA-RL & yes & supervised fit to critic-optimized samples & the samples start from its own policy (own samples) \\
DPPO & last denoising steps & on-policy gradient over the denoising MDP & PPO clip (previous iterate) \\
ReinFlow & yes & on-policy gradient over the noise-injected chain & PPO clip (previous iterate) \\
FPO & yes & on-policy; flow matching loss ratio in the PPO clip & PPO clip (previous iterate) \\
SAC Flow & yes, through the chain & direct $Q$-gradient through the reparameterized chain & $\beta\,\|a_\theta - a\|^2$ to replay actions when fine-tuning (data) \\
Q-Flow & yes & velocity matching to a flow matching target plus the value gradient & the flow matching target (data); offline \\
Q-VGM & yes & velocity matching to a normalized, gated $Q$-gradient step from its own endpoint & no separate term; the step is bounded (own endpoint) \\
TRQAM & yes & adjoint matching & adaptive path-space KL (pretrained policy) \\
FMQ & yes & regression onto a $Q$-guided target & radius $\eta$ around the reference policy (pretrained policy) \\
OFQL & yes, one-step & direct $Q$-gradient & MeanFlow loss on the data (data); offline \\
MeanFlowQL & yes, one-step & direct $Q$-gradient & MeanFlow loss on the data (data) \\
DICE-RL & no (frozen); a residual & direct $Q$-gradient into the residual & selective penalty $\|s_\theta\|^2$ on the residual (pretrained policy) \\
SWFP & appended flow blocks & soft-$Q$ energy with entropy & $W_2$ penalty (previous iterate) \\
TD3-C & yes (deterministic) & direct $Q$-gradient & $\ell_2$ bound to the target policy (previous iterate) \\
PROTO & yes (Gaussian) & direct $Q$-gradient & KL penalty (previous iterate) \\
SFG (ours) & yes, full decoder & direct $Q$-gradient & none; a bound on the functional step (previous committed actor) \\
\bottomrule
\end{tabular}
\caption{The methods cited in Section~\ref{sec:related} on the axes of Table~\ref{tab:baselines}, with the path by which the critic's signal reaches the policy. Each row follows the method's own paper; the reference of the anchor or constraint is in parentheses. The methods of Table~\ref{tab:baselines} are not repeated. Among the methods that train the pretrained policy's own weights, only Q-VGM and SFG add no term toward the data or the pretrained policy; Q-VGM bounds a per-update step of a surrogate target, SFG bounds the per-episode change of the committed actor.}
\label{tab:related_axes}
\end{table}

%% file: sections/table_main_full_v2_body.tex
\begin{tabular}{@{}l ccccccc c@{}}
\toprule
& square & tool\_hang & k-mixed & k-partial & k-complete & \rev{can} & \rev{lift} & \rev{Mean} \\
\midrule
\multicolumn{9}{@{}l}{\emph{Task completion score at 1.4M steps (mean of the evaluations in the last 200k)}} \\
DSRL & 0.76 & 0.01 & 0.45 & 0.57 & 0.52 & \rev{0.84} & \rev{\textbf{1.00}} & \rev{0.59} \\
EXPO & 0.83 & 0.24 & 0.52 & 0.77 & 0.05 & \rev{0.97} & \rev{\textbf{1.00}} & \rev{0.63} \\
QC-FQL & 0.39 & 0.00 & 0.96 & 0.92 & 0.85 & \rev{0.61} & \rev{\textbf{1.00}} & \rev{0.68} \\
DFP & 0.97 & 0.42 & 0.95 & 0.98 & 0.92 & \rev{0.89} & \rev{\textbf{1.00}} & \rev{0.88} \\
MVP & 0.98 & 0.67 & 0.99 & 0.97 & \textbf{1.00} & \rev{0.88} & \rev{\textbf{1.00}} & \rev{0.93} \\
OGPO+CA & 0.84 & 0.90 & 0.87 & 0.87 & 0.99 & \rev{0.86} & \rev{\textbf{1.00}} & \rev{0.91} \\
\rev{SFG (10-step flow)} & \rev{\textbf{0.99}} & \rev{0.66} & \rev{0.87} & \rev{0.93} & \rev{0.98} & \rev{\textbf{1.00}} & \rev{\textbf{1.00}} & \rev{0.92} \\
\rev{SFG (1-step flow)} & \rev{\textbf{0.99}} & \rev{\textbf{0.91}} & \rev{0.91} & \rev{0.87} & \rev{0.93} & \rev{\textbf{1.00}} & \rev{\textbf{1.00}} & \rev{0.94} \\
SFG (MeanFlow) & 0.98 & \textbf{0.91} & \textbf{1.00} & \textbf{1.00} & \textbf{1.00} & \rev{\textbf{1.00}} & \rev{\textbf{1.00}} & \rev{\textbf{0.98}} \\
\midrule
\multicolumn{9}{@{}l}{\emph{Environment steps (M) at which the smoothed score first reaches 0.8}} \\
DSRL & -- & -- & -- & -- & -- & \rev{1.10} & \rev{0.14} & \rev{1.18} \\
EXPO & 1.24 & -- & -- & -- & -- & \rev{\textbf{0.00}} & \rev{\textbf{0.00}} & \rev{0.98} \\
QC-FQL & -- & -- & \textbf{0.22} & 0.38 & 0.60 & \rev{--} & \rev{\textbf{0.00}} & \rev{0.77} \\
DFP & 0.36 & -- & 1.14 & 1.08 & 1.22 & \rev{0.14} & \rev{\textbf{0.00}} & \rev{0.76} \\
MVP & 0.28 & -- & 0.78 & 1.04 & 0.56 & \rev{0.32} & \rev{0.12} & \rev{0.64} \\
OGPO+CA & 1.04 & 0.61 & 0.78 & 0.78 & \textbf{0.01} & \rev{0.72} & \rev{\textbf{0.00}} & \rev{0.56} \\
\rev{SFG (10-step flow)} & \rev{0.34} & \rev{--} & \rev{1.14} & \rev{0.82} & \rev{0.56} & \rev{0.34} & \rev{0.14} & \rev{0.68} \\
\rev{SFG (1-step flow)} & \rev{0.24} & \rev{0.66} & \rev{0.94} & \rev{0.80} & \rev{0.36} & \rev{0.22} & \rev{0.04} & \rev{0.47} \\
SFG (MeanFlow) & \textbf{0.18} & \textbf{0.28} & 0.38 & \textbf{0.26} & 0.28 & \rev{0.06} & \rev{\textbf{0.00}} & \rev{\textbf{0.21}} \\
\bottomrule
\end{tabular}

%% file: sections/appendix_trust_region.tex
\section{Relation to Trust-Region and Penalty Methods}
\label{app:trust}

The functional step projection is a trust region on the generated actions: once per episode it bounds the root-mean-square change of the actions that a deterministic generator produces on a fixed bank of state--noise pairs~\eqref{eq:fstep}, measured relative to the previous committed actor, so that the reference advances during fine-tuning. It is enforced by a projection at the episode boundary rather than by a penalty or a clipped ratio inside the loss: within the episode the critic's gradient is followed unmodified, and at the boundary the accumulated update is rescaled to the bound, with no coefficient to balance against the objective. TRPO and PPO bound a divergence between consecutive policy distributions~\citep{schulman2015trpo,schulman2017ppo}, as do SWFP and MPO~\citep{sun2025swfp,abdolmaleki2018mpo}; a one-step generator has no tractable density, and the change of its outputs on a fixed probe bank is the analog that a forward pass can measure. The value of the bound is chosen empirically (Section~\ref{sec:method_cap}).

Penalties toward the data or the pretrained policy, including the trust regions that FMQ and TRQAM place around the pretrained flow policy~\citep{ziakas2026fmq,dong2026trqam} and the proximity terms of some flow-policy methods~\citep{zhang2026sacflow}, are anchors in the sense of Section~\ref{sec:prelim_rl}: they constrain where the policy may end. TD3-C and PROTO~\citep{luo2023td3c,li2023proto} bound the step from the previous policy for Gaussian actors and are the closer relatives, with the divergence replaced here by the functional step. Section~\ref{sec:analysis} motivates the bound by the loss of pretrained competence within the first updates; whether it also limits the exploitation of critic errors is not separated by our experiments.

\subsection{\rev{Empirical comparison with a penalty and a parameter-space trust region}}
\label{app:trust_empirical}
\rev{\textbf{Penalty.} We add $\lambda \lVert a_\theta(s,z) - a_{\mathrm{ref}}(s,z) \rVert^2$ on the batch states to the actor loss at every gradient step, without the projection, where $a_{\mathrm{ref}}$ is the actor at the start of the episode. $\lambda = 4$, calibrated on \texttt{tool\_hang} so that the committed functional step at the start of training matches $\delta = 0.01$, is also used on \texttt{kitchen-partial}; on \texttt{square} the calibrated value is $3.3$.}

\rev{\textbf{Trust region.} At the end of each episode, if the parameter distance $n = \lVert \theta - \theta_{\mathrm{ref}} \rVert_2$ exceeds $\epsilon$, we commit $\theta_{\mathrm{ref}} + (\epsilon / n)(\theta - \theta_{\mathrm{ref}})$ instead of $\theta$; only the metric differs from SFG. $\epsilon$ is calibrated on each task to the same starting step ($2.57$, $6.35$, and $5.98 \times 10^{-3}$). Everything else follows the SFG runs (pretrained policy, critic, replay, learning rate, evaluation).}

\begin{figure}[H]
\centering
\includegraphics[width=\linewidth]{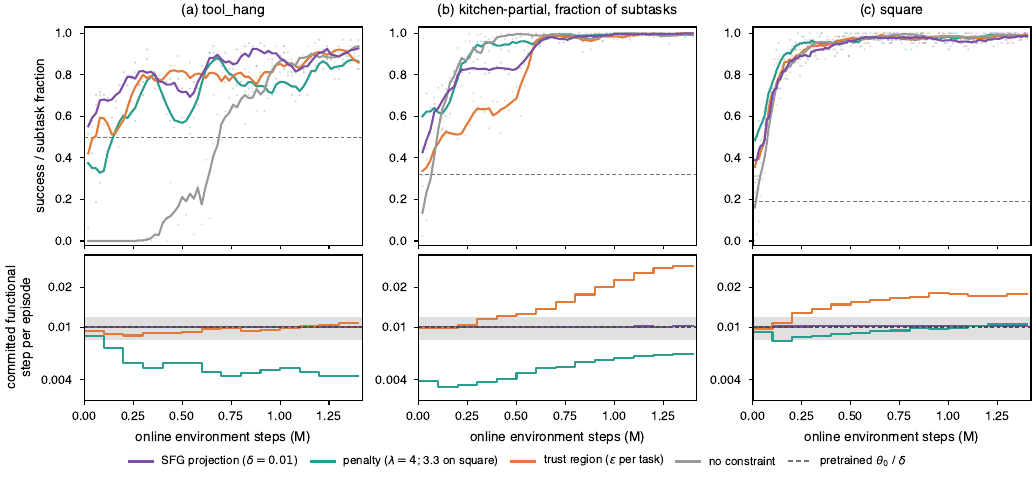}
\caption{\rev{The runs of Table~\ref{tab:constraints}. Top: evaluations as dots and their centered moving average over $\pm$50k steps as lines. Bottom: median committed functional step per episode in each 100k-step bin, on a log scale; the dashed line is $\delta = 0.01$ and the grey band $\pm 20\%$. The unconstrained runs have no committed step.}}
\label{fig:constraints}
\end{figure}

\rev{Both constraints avoid the early collapse of the unconstrained update (Table~\ref{tab:constraints}, Figure~\ref{fig:constraints}). The trust region is the weaker one: the same distance in parameter space corresponds to very different changes in the generated actions across pretrained networks, so $\epsilon$ has to be tuned for each task, and its functional step still drifts during training, up to $2.9\delta$; averaged over the three tasks it is behind SFG (0.61 against 0.71 over the first 0.3M, 0.8 reached at 0.38M against 0.24M, area 0.67 against 0.75 over the first 0.5M). The penalty is a viable alternative and performs close to SFG on the three ablation tasks, faster on \texttt{kitchen-partial} and \texttt{square} and slower on \texttt{tool\_hang}. Averaged over the three tasks, SFG is slightly higher early and faster (0.71 against 0.68 over the first 0.3M, 0.8 reached at 0.24M against 0.35M), with the same area over the first 0.5M (0.75) and a similar final score (0.96 against 0.95).}

\begin{table}[H]
\centering
\footnotesize\setlength{\tabcolsep}{4.5pt}
\begin{tabular}{@{}l l c cc c c c@{}}
\toprule
& & committed step & \multicolumn{2}{c}{first 0.3M} & steps to & AUC & \\
\cmidrule(lr){4-5}
Arm & setting & over training & mean & lowest & 0.8 (M) & 0--0.5M & final \\
\midrule
\multicolumn{8}{@{}l}{\emph{\texttt{tool\_hang} ($\theta_0 = 0.50$)}} \\
SFG projection & $\delta = 0.01$ & 0.0100--0.0101 & 0.70 & 0.38 & 0.28 & 0.73 & 0.91 \\
penalty & $\lambda = 4$ & 0.0086 $\to$ 0.0043 & 0.51 & 0.19 & 0.66 & 0.58 & 0.86 \\
trust region & $\epsilon = 2.57 \times 10^{-3}$ & 0.0086--0.0108 & 0.59 & 0.06 & 0.40 & 0.67 & 0.90 \\
no constraint & -- & -- & 0.00 & 0.00 & 1.00 & 0.04 & 0.90 \\
\midrule
\multicolumn{8}{@{}l}{\emph{\texttt{kitchen-partial}, fraction of subtasks ($\theta_0 = 0.32$)}} \\
SFG projection & $\delta = 0.01$ & 0.0100--0.0101 & 0.70 & 0.32 & 0.26 & 0.71 & 1.00 \\
penalty & $\lambda = 4$ & 0.0035--0.0063 & 0.73 & 0.45 & 0.24 & 0.81 & 1.00 \\
trust region & $\epsilon = 6.35 \times 10^{-3}$ & 0.0099 $\to$ 0.0287 & 0.50 & 0.27 & 0.56 & 0.52 & 1.00 \\
no constraint & -- & -- & 0.65 & 0.02 & 0.22 & 0.76 & 0.99 \\
\midrule
\multicolumn{8}{@{}l}{\emph{\texttt{square} ($\theta_0 = 0.19$)}} \\
SFG projection & $\delta = 0.01$ & 0.0100--0.0102 & 0.73 & 0.30 & 0.18 & 0.81 & 0.98 \\
penalty & $\lambda = 3.3$ & 0.0079--0.0106 & 0.81 & 0.28 & 0.14 & 0.86 & 0.99 \\
trust region & $\epsilon = 5.98 \times 10^{-3}$ & 0.0096 $\to$ 0.0177 & 0.74 & 0.23 & 0.18 & 0.82 & 0.99 \\
no constraint & -- & -- & 0.69 & 0.00 & 0.18 & 0.79 & 0.99 \\
\midrule
\multicolumn{8}{@{}l}{\emph{Mean of the three tasks, one setting per method}} \\
SFG projection & & & 0.71 & 0.33 & 0.24 & 0.75 & 0.96 \\
penalty & & & 0.68 & 0.31 & 0.35 & 0.75 & 0.95 \\
trust region & & & 0.61 & 0.19 & 0.38 & 0.67 & 0.96 \\
no constraint & & & 0.45 & 0.01 & 0.47 & 0.53 & 0.96 \\
\bottomrule
\end{tabular}
\caption{\rev{The projection against a penalty and a parameter-space trust region. ``Committed step'': range (or first $\to$ last value) of the per-100k median of the committed functional step per episode on the probe bank, target 0.01. ``First 0.3M'': mean and lowest evaluation (\texttt{square}, evaluated every 10k steps: all evaluations up to 0.3M). ``Steps to 0.8'': as in Table~\ref{tab:main}. ``AUC 0--0.5M'': area under the evaluation curve over the first 0.5M steps, with $\theta_0$ at step 0, divided by 0.5M. ``Final'': mean of the evaluations at 1.2M steps and later.}}
\label{tab:constraints}
\end{table}

